\documentclass[runningheads]{llncs}

\usepackage{eccv}

\usepackage{eccvabbrv}

\usepackage{graphicx}
\usepackage{booktabs}
\usepackage{array,multirow}
\usepackage{diagbox}
\usepackage{algorithm}
\usepackage{algorithmic}
\usepackage[normalem]{ulem}
\usepackage{colortbl}
\useunder{\uline}{\ul}{}

\definecolor{baselinecolor}{gray}{.9}
\newcommand{\baseline}[1]{\cellcolor{baselinecolor}{#1}}

\usepackage[accsupp]{axessibility}  

\usepackage[pagebackref,breaklinks,colorlinks,citecolor=eccvblue]{hyperref}

\usepackage{amsmath,amsfonts,bm}

\def\eqref#1{equation~\ref{#1}}

\def\1{\bm{1}}

\def\vq{{\bm{q}}}

\def\vy{{\bm{y}}}

\def\mB{{\bm{B}}}

\def\mR{{\bm{R}}}

\DeclareMathAlphabet{\mathsfit}{\encodingdefault}{\sfdefault}{m}{sl}
\SetMathAlphabet{\mathsfit}{bold}{\encodingdefault}{\sfdefault}{bx}{n}
\newcommand{\tens}[1]{\bm{\mathsfit{#1}}}

\def\tF{{\tens{F}}}

\def\tW{{\tens{W}}}

\newcommand{\R}{\mathbb{R}}

\usepackage{orcidlink}

\makeatletter
\def\@fnsymbol#1{\ensuremath{\ifcase#1\or \dagger \or\faEnvelope[regular]\fi}}
\makeatother

\begin{document}

\title{BKDSNN: Enhancing the Performance of Learning-based Spiking Neural Networks Training with Blurred Knowledge Distillation} 

\titlerunning{Enhancing SNN with Blurred Knowledge Distillation}

\author{Zekai Xu\inst{1}\and
Kang You\inst{1} \and
Qinghai Guo\inst{2} \and
Xiang Wang\inst{2} \and
Zhezhi He\inst{1}\thanks{Corresponding Author}\orcidlink{0000-0002-6357-236X}
}

\authorrunning{Zekai Xu et al.}

\institute{School of Electronic Information and
Electrical Engineering, Shanghai Jiao Tong University, Shanghai, China \and
Huawei Technologies, Shenzhen, China\\
\email{\{sherlock.holmes.xu,kang\_you,\textsuperscript{$\dagger$}zhezhi.he\}@sjtu.edu.cn}   
\url{https://github.com/Intelligent-Computing-Research-Group/BKDSNN}
}

\maketitle

\begin{abstract}
  Spiking neural networks (SNNs), which mimic biological neural systems to convey information via discrete spikes, are well-known as brain-inspired models with excellent computing efficiency. By utilizing the surrogate gradient estimation for discrete spikes, learning-based SNN training methods that can achieve ultra-low inference latency (\ie, number of time-step) have emerged recently. Nevertheless, due to the difficulty of deriving precise gradient for discrete spikes in learning-based methods, a distinct accuracy gap persists between SNNs and their artificial neural networks (ANNs) counterparts. To address the aforementioned issue, we propose a \emph{blurred knowledge distillation (BKD)} technique, which leverages randomly blurred SNN features to restore and imitate the ANN features. Note that, our BKD is applied upon the feature map right before the last layer of SNNs, which can also mix with prior logits-based knowledge distillation for maximal accuracy boost. In the category of learning-based methods, our work achieves state-of-the-art performance for training SNNs on both static and neuromorphic datasets. On the ImageNet dataset, BKDSNN outperforms prior best results by 4.51\% and 0.93\% with the network topology of CNN and Transformer, respectively. 
\end{abstract}    
\section{Introduction}
\label{sec:intro}

Inspired by the neural firing mechanism in the biological nervous system, spiking neural network (SNN)~\cite{maass1997networks} emerges as an alternative framework for the traditional artificial neural network (ANN). ANNs rely on continuous activation values to convey information between neurons~\cite{lecun2015deep}, while SNNs utilize discrete spiking events to process and propagate information~\cite{davies2018loihi, merolla2014million}. Given the remarkable energy efficiency achieved by such event-driven information propagation~\cite{han2020rmp}, SNNs show promising prospects on computational intelligence tasks~\cite{roy2019towards} with strong autonomous learning capabilities and ultra-low power consumption~\cite{bu2023optimal, ding2022snn, ostojic2014two, zenke2015diverse}. Apart from these advantages, SNNs are also natively compatible with event-based neuromorphic datasets~(\eg, CIFAR10-DVS)~\cite{fang2021deep, zhou2022spikformer, zhou2023spikingformer, zhou2023enhancing}.

Currently, \textit{conversion-based} and \textit{learning-based} are two common methods to train SNNs. Conversion-based methods attempt to leverage information from ANNs by transferring the parameters of a pre-trained ANN to its SNN counterpart, which have achieved promising results in several tasks~\cite{bu2023optimal, deng2021optimal, li2022quantization, li2021free}. However, event-triggered neurons convey limited information with ultra-low inference time-steps (\eg, 4 time-steps), leading to degraded performance of conversion-based methods \cite{fang2021deep, zhou2022spikformer}. Consequently, conversion-based methods usually take a long inference time~(\eg, 32 time-steps) to get comparable accuracy as the original ANN~\cite{cao2015spiking, sengupta2019going} based on the equivalence of the ReLU activation and the firing rate of integrate-and-fire (IF) neuron~\cite{gerstner2002spiking}. In order to further improve the performance of SNNs with ultra-low inference time-steps, learning-based methods unfold SNN training over the inference time-steps and leverage back-propagation through time (known as BPTT)~\cite{zenke2021remarkable, wu2019direct} to train SNNs. Inspired by~\cite{kheradpisheh2018stdp}, several works~\cite{liu2021sstdp, fang2021deep, zhou2023enhancing} adopt surrogate gradient estimation as an implicit differentiation for direct training of SNNs. With the combination of BPTT and surrogate gradient estimation, learning-based methods exhibit advantages in processing event-based neuromorphic datasets~(\eg, CIFAR10-DVS~\cite{2017CIFAR10})~\cite{fang2021deep, zhou2022spikformer, zhou2023spikingformer, zhou2023enhancing}. Furthermore, direct training of CNN- \cite{hu2020spiking, fang2021deep} and Transformer-based \cite{zhou2022spikformer, zhou2023spikingformer, zhou2023enhancing} SNNs also achieves promising performance with ultra-low time-steps on static datasets~(\eg, CIFAR10~\cite{krizhevsky2009learning}, CIFAR100~\cite{krizhevsky2009learning} and ImageNet~\cite{deng2009imagenet}). 

Despite that numerous efforts have been made in learning-based methods, SNNs still need improvement compared with ANNs. ~\cite{zhou2023enhancing, xu2023constructing} claim that imprecise surrogate gradient estimation of spiking neurons leads to performance degradation. On the other hand, KDSNN~\cite{xu2023constructing} and LaSNN~\cite{hong2023lasnn} introduce layer-wise knowledge distillation~(KD) to train a student SNN with rich information from the ANN teacher, achieving promising performance on several classical vision datasets~(MNIST, CIFAR10/100) with CNN-based models. However, according to our preliminary experimental results, both KDSNN and LaSNN struggle with more complex datasets~(ImageNet) and Transformer-based models. We hypothesize that the direct distillation between discrete SNN features and continuous ANN features~\cite{xu2023constructing, hong2023lasnn} might be the source of the limitations.  


In this paper, to further alleviate performance degradation of learning-based SNNs, we introduce a blurred knowledge distillation~(BKD) technique to leverage randomly blurred SNN features to restore and mimic ANN features. We apply BKD to the intermediate feature before the last layer of the student SNN and prove its superiority over layer-wise knowledge distillation in learning-based SNNs. Furthermore, we combine prior logits-based distillation with BKD, which provides mutual benefits for improving performance. We evaluate BKDSNN on both static datasets~(\ie, CIFAR10, CIFAR100, and ImageNet) and a neuromorphic dataset~(\ie, CIFAR10-DVS), where we achieve state-of-the-art (SOTA) performance with both CNN- and Transformer-based SNN models. 


To summarize, our contributions are threefold:
\begin{itemize}
    \item[$\bullet$] We introduce a blurred knowledge distillation~(BKD) technique to guide the training of student SNNs through feature imitation and demonstrate the advantages of BKD on learning-based SNNs.
    \item[$\bullet$] We apply BKD to the intermediate feature before the last layer and prove its superiority over previous layer-wise distillation. Combined with prior logits-based knowledge distillation, our BKDSNN maximizes accuracy boost on learning-based SNNs.
    \item[$\bullet$] We benchmark our method on both static datasets~(\ie, CIFAR10, CIFAR100 and ImageNet) and a neuromorphic dataset~(\ie, CIFAR10-DVS). Our method outperforms prior SOTA learning-based methods for both CNN- and Transformer-based models.
\end{itemize}

\section{Background and Related Work}
\label{sec:formatting}


\paragraph{Spiking Neuron Model.}

Previous works~\cite{cao2015spiking, deng2021optimal, fang2021deep, zhou2022spikformer, zhou2023spikingformer, zhou2023enhancing, bu2023optimal, hu2023fast} have widely used the classic integrate-and-fire (IF) neuron as the fundamental unit in SNNs. IF neuron integrates input currents through changes in its membrane potential, which is compared against the threshold for spike generation. The membrane potential is reset after emitting a spike. The dynamics of the IF neuron is depicted as follows:
\begin{equation}
\label{eq:neuron}
\begin{gathered}
H[t]=V[t-1]+\frac{1}{\tau}\left(I[t]-\left(V[t-1]-V_{\text{reset}}\right)\right) \\
S[t]=\Theta\left(H[t]-V_{\text{th}}\right)= \begin{cases}1, & H[t] \geq V_{\text {th }} \\ 0, & H[t]<V_{\text {th }}\end{cases} \\
V[t]=H[t](1-S[t])+V_{\text {reset }} S[t]
\end{gathered}
\end{equation}
where $\tau$ is the membrane time constant, $I[t]$ is the receiving current at time step $t$ and $\Theta\left(\cdot\right)$ is the heaviside function. The neuron accumulates $I[t]$ to change the membrane potential $H[t]$ at time step $t$. Once $H[t]$ exceeds the firing threshold $V_{\text{th}}$, the neuron generates a spike and resets the membrane potential to $V_{\text{reset}}$.



\paragraph{Conversion-based Methods.}

To fully make use of the knowledge from pre-trained ANN, conversion-based methods are designed to transfer the learned parameters of ANNs into corresponding SNNs, without further training of SNNs. Previous works~\cite{han2020deep, han2020rmp} replace ReLU activations in ANNs with spiking neurons and add scaling operations like weight normalization and threshold balancing~\cite{diehl2015fast} to generate corresponding SNNs. Although several works~\cite{deng2021optimal, han2020deep, han2020rmp, li2021free} have achieved nearly loss-less accuracy with VGG-16 and ResNet, the converted SNNs still require long inference time to match the original ANNs in precision. \cite{rueckauer2017conversion} claims that low firing rates caused by weight normalization result in increasing latency for the information reaching higher layers, which limits the practical application of the converted SNNs.


\begin{figure*}[t]
    \centering
	\subcaptionbox{Architecture of BKDSNN.\label{fig:arch}}{\includegraphics[width=0.68\linewidth]{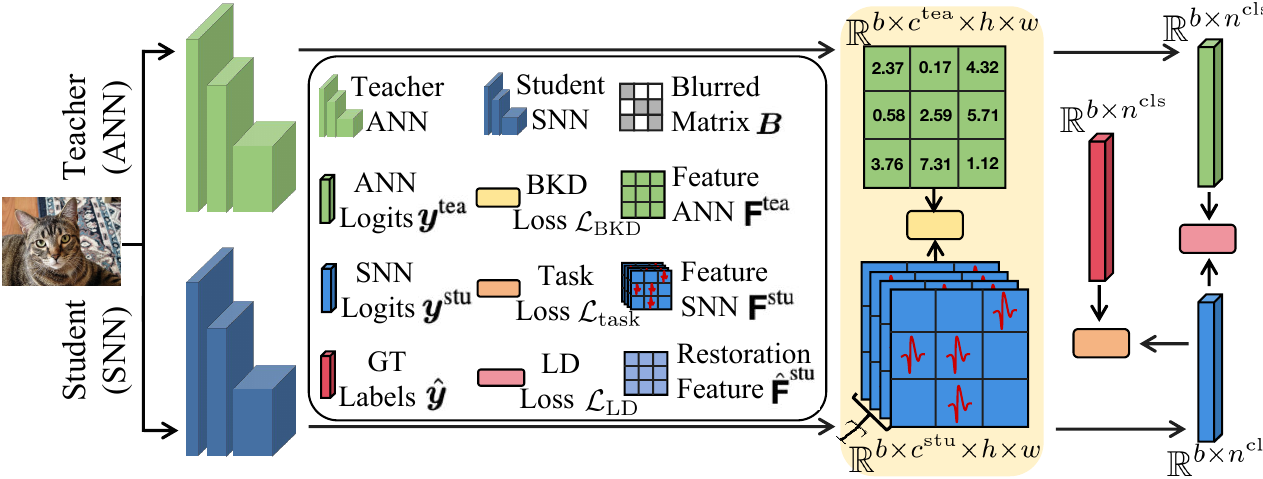}}
	\subcaptionbox{BKD block.\label{fig:bkd}}
   {\includegraphics[width=0.31\linewidth]{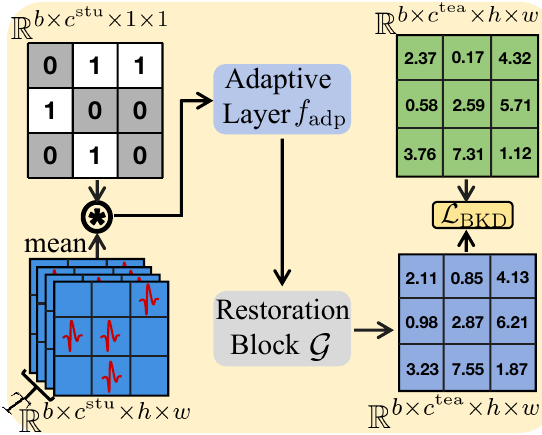}}
	\caption{\textbf{Overview of training SNN with blurred knowledge distillation (BKD)}. The SNN is directly trained with back-propagation through time (BPTT)~\cite{wu2019direct}, where we utilize a blurred variant of KD (\ie, BKD) to achieve higher accuracy. BKD highlighted in yellow shaded region differs from prior SNN-oriented KD in three perspectives: 1) A blurred matrix $\mB$ is \emph{randomly sampled on the fly} (per input image) to mask out the feature of student SNN; 2) A restoration block $\mathcal{G}$ consisting of two convolutional layers connected by a ReLU layer is applied on \emph{blurred SNN features} to \emph{restore and mimic ANN features}; 3) Such blurred knowledge distillation is applied only to the \emph{intermediate features before the last layer}.}
	\label{fig:architecture}
 \vskip -0.5cm
\end{figure*}

\paragraph{Learning-based Methods.}

With the development of surrogate gradient estimation~\cite{neftci2019surrogate}, learning-based methods leverage continuous-value-based loss functions to handle nondifferentiable difficulties in SNN training. Ref. \cite{neftci2019surrogate} introduces surrogate gradient estimation to conduct SNN training on the event-based DVS gesture dataset, which achieves better accuracy after a few training iterations. In an effort to improve computational resources in SNNs, joint CNN-SNN~\cite{xu2022hierarchical} puts forward a novel augmented spiking-based framework and achieves promising results in several image classification tasks. DCT-SNN~\cite{garg2021dct} reduces the number of inference time-steps by combining frequency-domain learning with surrogate gradient descent. Spiking ResNet~\cite{zheng2021going} uses the directly-trained deep SNN to achieve high performance on ImageNet. SEW ResNet~\cite{fang2021deep} proposes the spike-element-wise block to counter gradient vanishing and gradient exploding, which successfully extends the depth of directly-trained SNN to more than 100 layers. When it comes to Transformer-based SNN, Spikformer~\cite{zhou2022spikformer} with a novel spiking self-attention block achieves significant improvement in image classification on both neuromorphic and static datasets. 
Transformer-based SNN is further improved by successive Spikingformer~\cite{zhou2023spikingformer}, ConvBN-MaxPooling-LIF~(CML)~\cite{zhou2023enhancing} and Meta-SpikeFormer~\cite{yao2023spike}, achieving SOTA performance.


\paragraph{Knowledge Distillation for SNN Training.}

Knowledge distillation (KD) is a classical transfer learning method that has been proven to be effective on various tasks, such as model compression~\cite{hinton2015distilling, sun2019patient, liu2019structured, yang2022focal}, incremental learning~\cite{peng2021sid, chen2019new, cermelli2020modeling, xu2022delving}, \etc. Recently, several works~\cite{lee2021energy, takuya2021training, kushawaha2021distilling} extend KD for SNN training, such as distilling spikes from a large SNN to a small SNN or logits-based knowledge from a well-trained ANN to a simple SNN. KDSNN~\cite{xu2023constructing} proposes a novel framework embracing both logits-based and feature-based knowledge distillation, which enhances performance on several image classification datasets~(MNIST, CIFAR10, and CIFAR100). LaSNN~\cite{hong2023lasnn} puts forward a layer-wise feature-based ANN-to-SNN distillation framework, achieving comparable top-1 accuracy to ANNs on the challenging Tiny ImageNet dataset. Furthermore, KDSNN~\cite{xu2023constructing} and LaSNN~\cite{hong2023lasnn} are applicable to both homogeneous and heterogeneous networks. Despite the superiority of KDSNN~\cite{xu2023constructing} and LaSNN~\cite{hong2023lasnn}, they struggle with more complex datasets~(\eg, ImageNet) with CNN-based and Transformer-based models according to our preliminary experimental results in~\cref{sec:exp-imagenet}. 

\section{Method}
\label{sec:method}

In this paper, we propose a novel blurred knowledge distillation SNN~(overall architecture of BKDSNN is in~\cref{fig:arch}) training method that leverages knowledge from the teacher ANN to guide the training of the student SNN. As depicted in~\cref{fig:bkd}, our BKD first introduces a randomly blurred matrix $\mB$ to mask the student SNN features. Unlike KDSNN~\cite{xu2023constructing} and LaSNN~\cite{hong2023lasnn}, which only apply an adaptive layer to align the SNN feature with the ANN feature before distillation, BKD introduces a restoration block $\mathcal{G}$ on blurred SNN feature to restore and mimic ANN feature. The proposed BKD is applied only to the intermediate feature before the last layer, different from the layer-wise distillation~\cite{xu2023constructing, hong2023lasnn}. As shown in~\cref{fig:arch}, we combine the logits-based distillation with BKD to leverage both logits and feature knowledge from the teacher ANN.






\begin{table}[t]
\centering
\caption{Summary of notations in this paper}
\resizebox{0.7\linewidth}{!}{
\renewcommand{\arraystretch}{1.0}
\begin{tabular}{cl}
\toprule
Notation & \multicolumn{1}{c}{Description} \\ \midrule
$T$ & \begin{tabular}[c]{@{}l@{}}time-step\end{tabular} \\
$\lambda$ & \begin{tabular}[c]{@{}l@{}}blurred ratio\end{tabular} \\
$\mR \in \R^{b \times c^\text{tea}}$ & \begin{tabular}[c]{@{}l@{}}random matrix $\mR \sim U\left(0, 1\right)$\end{tabular} \\
$\mB \in \R^{b \times c^\text{tea}}$ & \begin{tabular}[c]{@{}l@{}}blurred matrix\end{tabular} \\
$\hat{\vy} \in \R^{b \times n^\text{cls}}$ & \begin{tabular}[c]{@{}l@{}}ground truth label\end{tabular} \\
$\vy^\text{stu}, \vy^\text{tea} \in \R^{b \times n^\text{cls}}$ & \begin{tabular}[c]{@{}l@{}}output logits of student SNN and teacher ANN\end{tabular} \\
$\vq^\text{stu}, \vq^\text{tea} \in \R^{b \times n^\text{cls}}$ & \begin{tabular}[c]{@{}l@{}}$\vy^\text{stu}$ and $\vy^\text{tea}$ after flattened processing\end{tabular} \\
$\tF^\text{tea} \in \R^{b \times c^\text{tea} \times h \times w}$ & \begin{tabular}[c]{@{}l@{}}feature before the last layer of teacher ANN\end{tabular} \\
$\tF^\text{stu} \in \R^{T \times b \times c^\text{stu} \times h \times w}$ & \begin{tabular}[c]{@{}l@{}}feature before the last layer of student SNN\end{tabular} \\
$\Tilde{\tF}^\text{stu} \in \R^{b \times c^\text{stu} \times h \times w}$ & \begin{tabular}[c]{@{}l@{}}average of $\tF^\text{stu}$ at time-step dimension~($T$)\end{tabular} \\
$f_\text{adp}$ & \begin{tabular}[c]{@{}l@{}}adaptive layer to align $\tF^\text{stu}$ with $\tF^\text{tea}$\end{tabular} \\
$\mathcal{G}$ & \begin{tabular}[c]{@{}l@{}}restoration block for $\tF^\text{stu}$\end{tabular} \\
$\hat{\tF}^\text{stu} \in \R^{b \times c^\text{tea} \times h \times w}$ & \begin{tabular}[c]{@{}l@{}}blurred restoration feature from $\tF^\text{stu}$\end{tabular} \\
\bottomrule
\end{tabular}
}
\label{tab:notation_table}
\vskip -0.5cm
\end{table}

\paragraph{Blurred Knowledge Distillation.}
To facilitate understanding, we present important notations in \cref{tab:notation_table}. The proposed blurred knowledge distillation is illustrated in \cref{fig:bkd}. We first introduce a random matrix $\mR$ which obeys uniform distribution between 0 and 1. Then we apply a blurred ratio $\lambda$ to $\mR$ to generate a blurred matrix $\mB$ through \cref{eq:bm}.
\begin{equation}
\mB_{i, j}= \begin{cases}0, & \text { if } \mR_{i, j}<\lambda \\ 1, & \text { Otherwise }\end{cases}
\label{eq:bm}
\end{equation}
For feature $\tF^\text{stu}$ at the last layer of student SNN, we first average $\tF^\text{stu}$ at time-step dimension ($T$) as:
\begin{equation}
\Tilde{\tF}^\text{stu}= \frac{1}{T} \displaystyle\sum_{i=1}^{T} \tF^\text{stu}_{i}
\label{eq:mean}
\end{equation}
After the averaging operation above, we apply a convolutional adaptive layer $f_\text{adp}$ to align $\Tilde{\tF}^\text{stu}$ with $\tF^\text{tea}_{i}$ when student feature dimension $c^\text{stu}$ mismatches with teacher feature dimension $c^\text{tea}$. After the alignment, we construct a restoration block $\mathcal{G}$ with two convolutional layers connected by a ReLU layer and apply it to $\Tilde{\tF}^\text{stu}$ with the blurred matrix $\mB$ to achieve blurred restoration feature $\hat{\tF}^\text{stu}$ as illustrated in \cref{eq:restoration}.
\begin{equation}
\hat{\tF}^\text{stu}= \begin{cases} \mathcal{G}\left(f_\text{adp}(\Tilde{\tF}^\text{stu})*\mB\right), & c^\text{stu} \neq c^\text{tea} \\
\mathcal{G}\left(\Tilde{\tF}^\text{stu}*\mB\right), & \text {otherwise}\end{cases}
\label{eq:restoration}
\end{equation}
We calculate blurred knowledge distillation loss $\mathcal{L}_\text{BKD}$ using quadratic norm ($L^{2}$) as follows.
\begin{equation}
\mathcal{L}_\text{BKD}= \displaystyle\sum_{i}^{b}\sum_{j}^{c^\text{stu}}\sum_{k}^{h}\sum_{l}^{w}\left(\hat{\tF}_{i, j, k, l}^\text{stu}-\tF_{i, j, k, l}^\text{tea}\right)^2
\end{equation}
We then combine blurred knowledge distillation loss $\mathcal{L}_\text{BKD}$ with the original cross entropy loss $\mathcal{L}_\text{CE}$ to get the total loss $\mathcal{L}_\text{total}$ as follows.
\begin{equation}
    \mathcal{L}_\text{total} = \mathcal{L}_\text{CE}(\vy^\text{stu}, \hat{\vy}) + \mathcal{L}_\text{BKD}
    \label{eq:tloss}
\end{equation}
\paragraph{Effectiveness of BKD.}
\begin{figure}[t]
    \centering
   \subcaptionbox{Feature visualization of ResNet-18.\label{fig:visfmapcop}}
   {\includegraphics[width=0.54\linewidth]{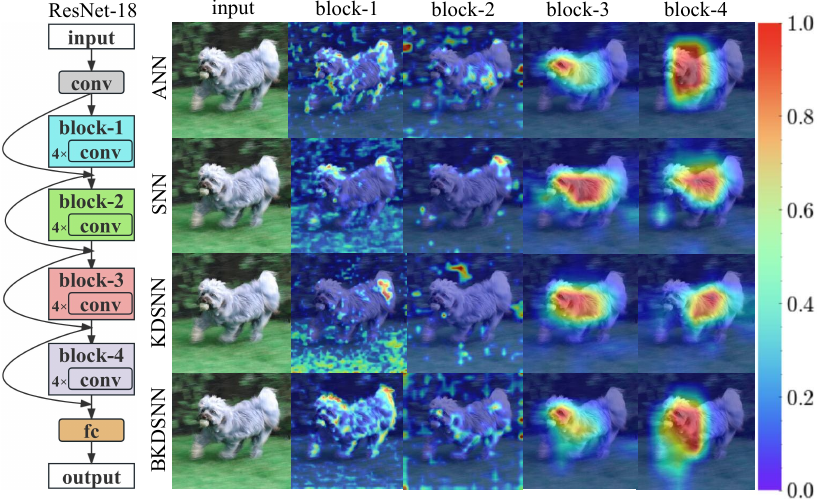}}
   \subcaptionbox{Feature distribution of SNN and ANN.\label{fig:fmapcompare}}
   {\includegraphics[width=0.45\linewidth]{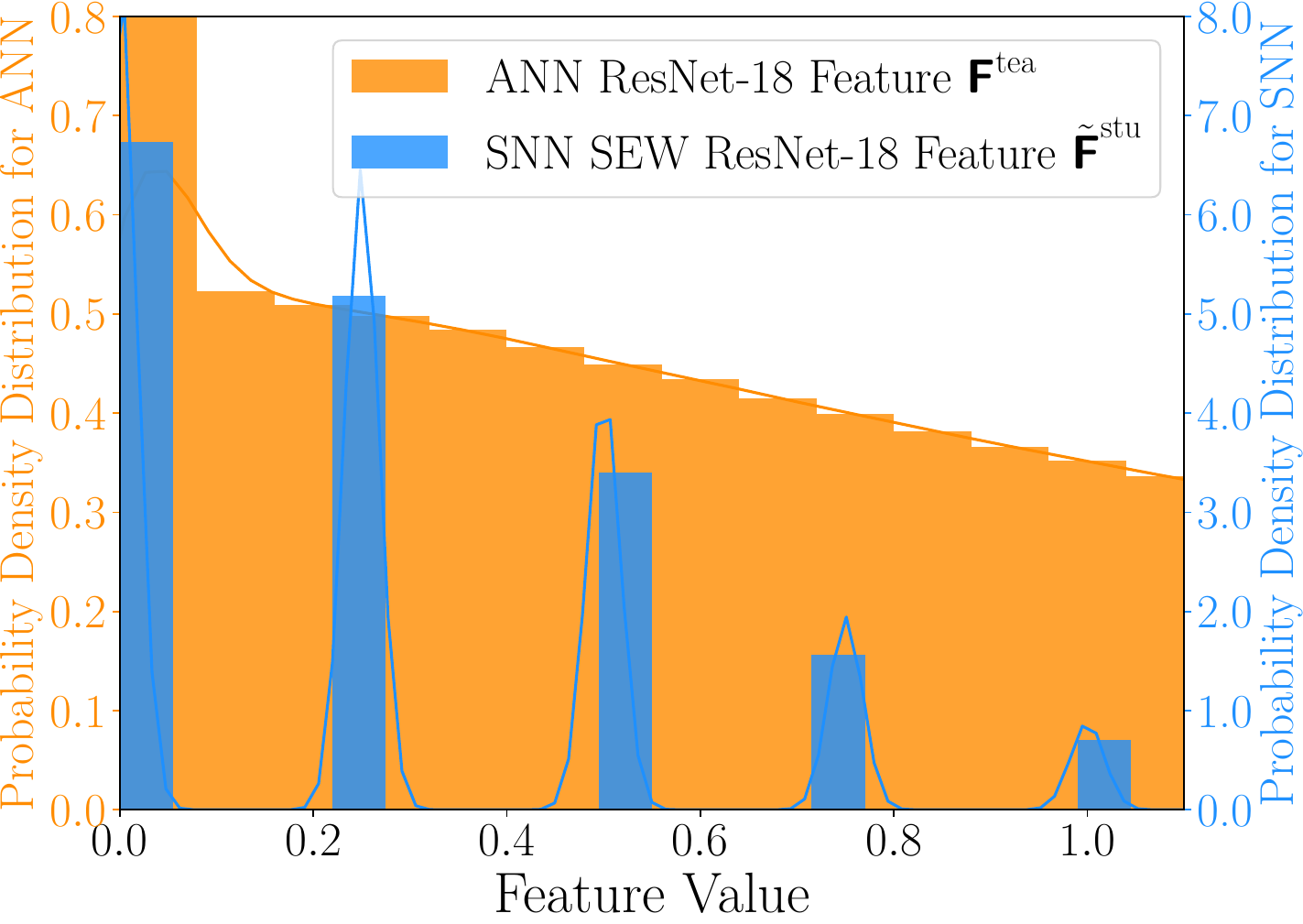}}
\caption{(a) Comparison of feature visualization in ResNet-18 under various methods; (b) Histogram of SNN and ANN feature.}
	\label{fig:bkdadv}
 \vskip -0.5cm
\end{figure}
For both CNN- and Transformer-based models, features of deeper layers have better representation for the original input image~\cite{khan2020survey, ranftl2021vision, zhou2021deepvit}. In other words, features of deeper layers contain sufficient information of adjacent pixels~(for CNN-based models)~\cite{khan2020survey} or patch pairs~(for Transformer-based models)~\cite{ranftl2021vision, zhou2021deepvit}. Therefore, we can use partial student SNN feature before the last layer to restore and mimic teacher ANN feature through a restoration block $\mathcal{G}$, which can help the student SNN achieves better representation and effectively mimics teacher ANN. We then apply Grad-CAM which can intuitively demonstrate the feature similarity~\cite{Selvaraju_2019} on each block in ResNet-18 under various methods. As depicted in~\cref{fig:visfmapcop}, compared with previous SNN~(SEW ResNet-18~\cite{fang2021deep}) and KDSNN~\cite{xu2023constructing}, our BKDSNN transfers sufficient knowledge from ANN and effectively mimics ANN feature. 
Meanwhile, as illustrated in \cref{fig:fmapcompare}, student SNN feature $\Tilde{\tF}^\text{stu}$ is discrete while teacher ANN feature $\tF^\text{tea}$ is continuous. Therefore, directly applying distillation to encourage discrete SNN feature $\Tilde{\tF}^\text{stu}$ to imitate continuous ANN feature $\tF^\text{tea}$~\cite{xu2023constructing, hong2023lasnn} is not plausible. However, the proposed BKD leverages the restoration block $\mathcal{G}$ to map discrete feature $\Tilde{\tF}^\text{stu}$ to continuous feature $\hat{\tF}^\text{stu}$ for imitating continuous ANN feature $\tF^\text{tea}$, which is a more proper spiking-friendly feature-based distillation method.

\paragraph{Mixed Distillation.} 

As depicted in \cref{fig:arch}, BKDSNN further applies logits-based distillation which transfers knowledge from teacher ANN logits to student SNN logits. As soft logits from the teacher model embrace sufficient hidden knowledge, which serves as a good regularization to student model~\cite{zhou2021rethinking}, we introduce the temperature $\tau$ to make the logits distribution flatter. The output logits of both teacher and student are processed as follows:
\begin{equation}
\vq_{i}=\frac{\exp \left(\vy_{i} / \tau\right)}{\textstyle\sum_{j=1}^{n^\text{cls}} \exp \left(\vy_{j} / \tau\right)}, \quad i \in \{1, 2, \dots , n^\text{cls}\}
\label{eq:flatten}
\end{equation}
Then logits-based distillation loss $\mathcal{L}_\text{LD}$ is defined as:
\begin{equation}
\mathcal{L}_\text{LD}=\tau^2 \cdot \mathcal{L}_\text { CrossEntropy }\left(\vq^\text{stu}, \vq^\text{tea}\right)
\label{eq:wsld}
\end{equation}
We combine $\mathcal{L}_\text{LD}$ with $\mathcal{L}_\text{BKD}$ to achieve mixed distillation loss $\mathcal{L}_\text{MD}$~(\cref{eq:md}), which contains both logits and feature information from the teacher ANN.
\begin{equation}
\mathcal{L}_\text{MD}= w_\text{LD} \cdot \mathcal{L}_\text{LD} + w_\text{BKD} \cdot \mathcal{L}_\text{BKD}
\label{eq:md}
\end{equation}
where $w_\text{LD}$ and $w_\text{BKD}$ are weights for $\mathcal{L}_\text{LD}$ and $\mathcal{L}_\text{BKD}$, respectively. Thus, the overall loss $\mathcal{L}$ is defined as follows:
\begin{equation}
\mathcal{L} = \mathcal{L}_\text { CrossEntropy }\left(\vy^\text{stu}, \hat{\vy}\right) + \mathcal{L}_\text{MD} = \mathcal{L}_\text { task} + \mathcal{L}_\text{MD}
\end{equation}
\vskip -0.3cm

\section{Experiments}
\label{sec:exp}

\subsection{Experimental Configuration}

\paragraph{Platform.}
All experiments are evaluated on a server equipped with 32 cores Intel Xeon Platinum 8352V CPU with 2.10GHz and 8-way NVIDIA GPUs. We use SpikingJelly~\cite{spikingjelly} to simulate IF neurons.

\paragraph{Static Datasets and Settings.}
We first evaluate BKDSNN on three static datasets, \ie, CIFAR10/100 and ImageNet.

\textit{CIFAR10/CIFAR100}~\cite{krizhevsky2009learning} contains 50k training and 10k testing images with 32$\times$32 resolution, and is labeled with 10- or 100-categories respectively for classification. 
For CNN-based models, we use ResNet-19 from TET~\cite{deng2022temporal} and lightweight ResNet-20 from SEW ResNet~\cite{fang2021deep} with corresponding ANNs as teacher models. 
For Transformer-based models, we train ViT-S/16 on CIFAR10 and CIFAR100 to make it serve as the teacher ANN model for Sformer-4-256, Sformer-2-384, and Sformer-4-384~(We abbreviate Spikingformer to Sformer for convenience). Sformer-L-D represents a Sformer model with L Sformer encoder blocks and D feature embedding dimensions.

\textit{ImageNet-1K}~\cite{deng2009imagenet} contains 1.3 millions of training images and 50k validation images with 224$\times$224 resolution, labeled with 1k classes. We choose ResNet-18, ResNet-34, and ResNet-50 pre-trained on ImageNet-1K as teacher ANN models for the corresponding SEW ResNet models~\cite{fang2021deep}. For Transformer-based models, we introduce ViT-B/16~\cite{dosovitskiy2020image} pre-trained on ImageNet1K as the teacher ANN model for Sformer-8-384, Sformer-8-512 and Sformer-8-768. We also apply our BKDSNN on the latest Meta-SpikeFormer~\cite{yao2023spike} and conduct experiments with Mformer-8-256, Mformer-8-384 and Mformer-8-512 on ImageNet-1K~(We abbreviate Meta-SpikeFormer to Mformer for convenience.).

For both CNN- and Transformer-based models, we set temperature $\tau$ to 2.0, blurred ratio $\lambda$ to 0.15 and logits-based distillation weight $w_\text{LD}$ to 1. \textbf{Specifically for CNN-based models}, we set blurred knowledge distillation weight $w_\text{BKD}$ to $7e^{-4}$ for CIFAR10/100 and $7e^{-5}$ for ImageNet. We follow all the training configurations (batch size, learning rate, optimizer, \etc) from SEW ResNet~\cite{fang2021deep} except ResNet-19, where we use training configuration from TET~\cite{deng2022temporal}. 
\textbf{For Transformer-based models}, we set $w_\text{BKD}$ to $2e^{-3}$ for both CIFAR10/100 and ImageNet, and we follow all the training configurations~(batch size, learning rate, optimizer, etc.) from Sformer+CML~\cite{zhou2023enhancing}.

\paragraph{Neuromorphic Dataset and Networks.}
We also evaluate BKDSNN on \textit{CIFAR10-DVS} \cite{2017CIFAR10}, which is a neuromorphic dataset that leverages dynamic vision sensor~(DVS) to convert 10k frame-based images from the CIFAR10 dataset into 10k event streams, providing an event-stream dataset with 10 different classes. It includes 9k training samples and 1k testing samples. Note that, CIFAR10-DVS is an event-stream dataset tailored to SNN, so we use SNN models with 16 time-steps as teacher models to guide the training of SNN models with lower time-steps~(\ie, 4 and 8). We utilize Wide-7B-Net~\cite{fang2021deep} and Sformer-2-256~\cite{zhou2023enhancing} with 4 and 8 inference time-steps as student models for CNN-based and Transformer-based models, respectively.

For both Wide-7B-Net~\cite{fang2021deep} and former-2-256~\cite{zhou2023enhancing}, we set $\tau$ to 2.0, $\lambda$ to 0.40 and $w_\text{LD}$ to 1. Then we set $w_\text{BKD}$ to $2e^{-3}$ for Wide-7B-Net~\cite{fang2021deep} and $7e^{-5}$ for Sformer-2-256~\cite{zhou2023enhancing}. We follow all the training configurations from SEW ResNet~\cite{fang2021deep} for Wide-7B-Net~\cite{fang2021deep}, and Sformer+CML~\cite{zhou2023enhancing} for Sformer-2-256~\cite{zhou2023enhancing}.

\paragraph{Settings for Prior KD Methods on SNNs.} Furthermore, to compare BKDSNN with previous KD methods on SNNs~(\ie, KDSNN~\cite{xu2023constructing} and LaSNN~\cite{hong2023lasnn}), we follow the settings from \cite{xu2023constructing} and \cite{hong2023lasnn} to reproduce the experimental results on all datasets with CNN-based and Transformer-based models.

\subsection{Comparison on Static Datasets}
\label{sec:exp-imagenet}

\paragraph{Comparison on CIFAR10/100} between BKDSNN and previous methods are listed in~\cref{tab:cifar-cnn} and~\cref{tab:cifar-transformer}, where BKDSNN surpasses prior SOTA methods with both CNN- and Transformer-based models. Specifically, BKD applied upon Sformer-4-384 outperforms previous SOTA by 0.18\% on CIFAR10 and 0.93\% on CIFAR100, which achieves 96.06\% and 81.26\% accuracy respectively. 

Intuitively, it is difficult for both KDSNN~\cite{xu2023constructing} and LaSNN~\cite{hong2023lasnn} to improve the performance of CNN-based models on the CIFAR dataset. 
For ResNet-19, the knowledge of the ANN teacher is limited for the SNN student, leading to the above difficulties. 
When it comes to ResNet-20, the lightweight model conveys limited information which results in the ineffectiveness of previous knowledge distillation methods. In contrast, our BKDSNN is effective on both limited teacher knowledge as well as lightweight student models, which verifies the superiority of our method.

\begin{table*}[t]
\caption{
\textbf{Comparison with previous works}, using various networks and multiple datasets. \textit{P} denotes \#parameters. \textit{KD} denotes knowldege distillation. $\star$: results taken from \cite{fang2021deep}. $\dagger$: prior state-of-the-art conversion-based method QCFS~\cite{bu2023optimal}. The prior SOTA results are marked in \colorbox{baselinecolor}{gray}, which is used to compare with our BKDSNN and report the accuracy difference. Best results are in \textbf{bold}. -- indicates results not available (not reported in original article or cannot reproduce).
}
\vskip -0.3cm
\begin{subtable}[c]{0.43\textwidth}
\centering
\resizebox{0.96\textwidth}{!}{
\begin{tabular}{clccccc}
\toprule
\multicolumn{2}{c}{\multirow{3}{*}{\diagbox[dir=SE,width=3cm,height=1cm]{Method}{Architecture}}} & \multicolumn{2}{c}{\multirow{2}{*}{\begin{tabular}[c]{@{}c@{}}\textit{R}-19\\ (\textit{P}: 12.63M)\end{tabular}}} & \multicolumn{2}{c}{\multirow{2}{*}{\begin{tabular}[c]{@{}c@{}}\textit{R}-20\\ (\textit{P}: 0.27M)\end{tabular}}} & \multirow{3}{*}{\begin{tabular}[c]{@{}c@{}}Time\\ Step\end{tabular}} \\
\multicolumn{2}{c}{}             & \multicolumn{2}{c}{}   & \multicolumn{2}{c}{}   &       \\
\multicolumn{2}{c}{}             & \begin{tabular}[c]{@{}c@{}}\textit{CF}10 \\ Acc\end{tabular}       & \begin{tabular}[c]{@{}c@{}}\textit{CF}100 \\ Acc\end{tabular}       & \begin{tabular}[c]{@{}c@{}}\textit{CF}10 \\ Acc\end{tabular}       & \begin{tabular}[c]{@{}c@{}}\textit{CF}100 \\ Acc\end{tabular}       &       \\ \midrule
\multirow{5}{*}{\rotatebox[origin=c]{90}{\parbox[c]{1.5cm}{\centering w/o \\ \textit{KD}}}}    & QCFS~\cite{bu2023optimal}$\dagger$            & -     & - &  83.75  &    55.23   & 4     \\
 & TET~\cite{deng2022temporal}            & \baseline{94.44}     & \baseline{74.47} &  87.75  &    59.14   & 4     \\
 & STBP~\cite{zheng2021going}      & 92.92          & 70.86 & --   & --   & 4     \\
 & Spiking-\textit{R}~\cite{hu2020spiking} &  92.13    &  70.13  &   87.32   &   59.93    & 4     \\
 & SEW-\textit{R}~\cite{fang2021deep}     &   93.24   &   70.84    &   \baseline{89.07}   &   60.16    & 4     \\ \midrule
 & Teacher ANN    &   95.60   &   79.78    &   91.77   &   68.40    & 1    \\ \midrule
\multirow{4}{*}{\rotatebox[origin=c]{90}{\parbox[c]{1cm}{\centering \centering with \textit{KD}}}} & KDSNN~\cite{xu2023constructing}          &   94.36   &   74.08    &   89.03   &    \baseline{60.18}   & 4     \\
 & LaSNN~\cite{hong2023lasnn}          &   94.21   &  73.92   &  88.41    &   59.17    & 4     \\
 & \begin{tabular}[l]{@{}c@{}}BKDSNN\\ (ours)\end{tabular}   &   {\begin{tabular}[c]{@{}c@{}}\textbf{94.64}\\ \textbf{(+0.20)}\end{tabular}}    &   \begin{tabular}[c]{@{}c@{}}\textbf{74.95}\\ \textbf{(+0.48)}\end{tabular}    &   \begin{tabular}[c]{@{}c@{}}\textbf{89.29}\\ \textbf{(+0.22)}\end{tabular}   &   \begin{tabular}[c]{@{}c@{}}\textbf{60.92}\\ \textbf{(+0.74)}\end{tabular}    & 4     \\ \bottomrule
\end{tabular}
}
\subcaption{CNN-based SNN on \textit{CF}10/100 dataset.\label{tab:cifar-cnn}}
\end{subtable}
\begin{subtable}[c]{0.57\textwidth}
\centering
\resizebox{0.99\textwidth}{!}{
\begin{tabular}{clclcclcclcc}
\toprule
\multicolumn{2}{c}{\multirow{4}{*}{\diagbox[dir=SE,width=3cm,height=1cm]{Method}{Architecture}}} & \multicolumn{3}{c}{\multirow{2}{*}{\begin{tabular}[c]{@{}c@{}}Sformer-4-256\\ (\textit{P}: 4.15M)\end{tabular}}}     & \multicolumn{3}{c}{\multirow{2}{*}{\begin{tabular}[c]{@{}c@{}}Sformer-2-384\\ (\textit{P}: 5.76M)\end{tabular}}}     & \multicolumn{3}{c}{\multirow{2}{*}{\begin{tabular}[c]{@{}c@{}}Sformer-4-384\\ (\textit{P}: 9.32M)\end{tabular}}}      & \multirow{3}{*}{\begin{tabular}[c]{@{}c@{}}Time\\ Step\end{tabular}} \\
\multicolumn{2}{c}{}              & \multicolumn{3}{c}{}            & \multicolumn{3}{c}{}            & \multicolumn{3}{c}{}            &   \\
\multicolumn{2}{c}{}              & \multicolumn{2}{c}{\begin{tabular}[c]{@{}c@{}}\textit{CF}10 \\ Acc\end{tabular}} & \begin{tabular}[c]{@{}c@{}}\textit{CF}100 \\ Acc\end{tabular}  & \multicolumn{2}{c}{\begin{tabular}[c]{@{}c@{}}\textit{CF}10 \\ Acc\end{tabular}} & \begin{tabular}[c]{@{}c@{}}\textit{CF}100 \\ Acc\end{tabular}  & \multicolumn{2}{c}{\begin{tabular}[c]{@{}c@{}}\textit{CF}10 \\ Acc\end{tabular}} & \begin{tabular}[c]{@{}c@{}}\textit{CF}100 \\ Acc\end{tabular}  &   \\ \midrule
\multirow{4}{*}{\rotatebox[origin=c]{90}{\parbox[c]{1.5cm}{\centering w/o \\ \textit{KD}}}}    & Spikformer~\cite{zhou2022spikformer} & \multicolumn{2}{c}{93.94} & 75.96 & \multicolumn{2}{c}{94.80} & 76.95 & \multicolumn{2}{c}{95.19} & 77.86 & 4 \\
 & +CML~\cite{zhou2023enhancing} & \multicolumn{2}{c}{94.82} & 77.64 & \multicolumn{2}{c}{95.63} & 78.75 & \multicolumn{2}{c}{95.93} & 79.65 & 4 \\
 & Sformer~\cite{zhou2023spikingformer}        & \multicolumn{2}{c}{94.77} & 77.43 & \multicolumn{2}{c}{95.22} & 78.34 & \multicolumn{2}{c}{95.61} & 79.09 & 4 \\& +CML~\cite{zhou2023enhancing}    & \multicolumn{2}{c}{94.94} & 78.19 & \multicolumn{2}{c}{95.54} & 78.87 & \multicolumn{2}{c}{95.81} & 79.98 & 4 \\ \midrule
 & Teacher ANN   & \multicolumn{2}{c}{96.75} & 82.22 & \multicolumn{2}{c}{96.75} & 82.22 & \multicolumn{2}{c}{96.75} & 82.22 & 1 \\ \midrule
\multirow{4}{*}{\rotatebox[origin=c]{90}{\parbox[c]{1cm}{\centering \centering with \textit{KD}}}} & KDSNN~\cite{xu2023constructing}          & \multicolumn{2}{c}{\baseline{95.00}} & \baseline{78.38} & \multicolumn{2}{c}{\baseline{95.59}} & \baseline{79.25} & \multicolumn{2}{c}{\baseline{95.88}} & \baseline{80.33} & 4 \\
 & LaSNN~\cite{hong2023lasnn}         & \multicolumn{2}{c}{94.97} & 78.02 & \multicolumn{2}{c}{95.55} & 78.91 & \multicolumn{2}{c}{95.79} & 79.99 & 4 \\
 & \begin{tabular}[l]{@{}c@{}}BKDSNN\\ (ours)\end{tabular} & \multicolumn{2}{c}{\begin{tabular}[c]{@{}c@{}}\textbf{95.29}\\ \textbf{(+0.29)}\end{tabular}}  & \begin{tabular}[c]{@{}c@{}}\textbf{79.41}\\ \textbf{(+1.03)}\end{tabular} & \multicolumn{2}{c}{\begin{tabular}[c]{@{}c@{}}\textbf{95.90}\\ \textbf{(+0.31)}\end{tabular}} & \begin{tabular}[c]{@{}c@{}}\textbf{80.63}\\ \textbf{(+1.38)}\end{tabular} & \multicolumn{2}{c}{\begin{tabular}[c]{@{}c@{}}\textbf{96.06}\\ \textbf{(+0.18)}\end{tabular}}  & \begin{tabular}[c]{@{}c@{}}\textbf{81.26}\\ \textbf{(+0.93)}\end{tabular} & 4 \\ \bottomrule
\end{tabular}
}
\subcaption{Transformer-based SNN on \textit{CF}10/100 dataset.\label{tab:cifar-transformer}}
\end{subtable}
\vspace{.5em}
\begin{subtable}[c]{0.48\textwidth}
\centering
\resizebox{0.96\textwidth}{!}{
\begin{tabular}{clcccc}
\toprule
\multicolumn{2}{c}{\multirow{2}{*}{\diagbox[dir=SE,width=3cm,height=0.8cm]{Method}{Architecture}}}         & \begin{tabular}[c]{@{}c@{}}\textit{R}-18\\ (\textit{P}: 11.69M)\end{tabular} & \begin{tabular}[c]{@{}c@{}}\textit{R}-34\\ (\textit{P}: 21.79M)\end{tabular} & \begin{tabular}[c]{@{}c@{}}\textit{R}-50\\ (\textit{P}: 25.56M)\end{tabular} & \multirow{2}{*}{\begin{tabular}[c]{@{}c@{}}Time\\ Step\end{tabular}} \\
\multicolumn{2}{c}{}         & Top1 Acc        & Top1 Acc         & Top1 Acc         &        \\ \midrule
   \multirow{5}{*}{\rotatebox[origin=c]{90}{\parbox[c]{1.5cm}{\centering w/o \\ \textit{KD}}}}    & QCFS~\cite{bu2023optimal}$\dagger$ & --   & 69.37 & --   & 32      \\
  & TET~\cite{deng2022temporal}     & --    & 68.00    & --    & 4      \\
  & STBP~\cite{zheng2021going}  & --    & 67.04 & --    & 4      \\
  & Spiking-\textit{R}~\cite{hu2020spiking}$\star$ & 62.32 & 61.86 & 57.66 & 4    \\
  & SEW-\textit{R}~\cite{fang2021deep} & 63.18           & 67.04 & 67.78 & 4      \\ \midrule
  & Teacher ANN & 69.76           & 73.31 & 76.31 & 1      \\ \midrule
\multirow{4}{*}{\rotatebox[origin=c]{90}{\parbox[c]{1cm}{\centering with \textit{KD}}}} & KDSNN~\cite{xu2023constructing}   &   \baseline{63.42}   &   \baseline{67.18}    &  67.72     & 4      \\
  & LaSNN~\cite{hong2023lasnn}   &  63.31    &    66.94   &   \baseline{67.81}    & 4      \\
  & \begin{tabular}[l]{@{}c@{}}BKDSNN\\ (ours)\end{tabular}  & \begin{tabular}[c]{@{}c@{}}\textbf{65.60}\\ \textbf{(+2.18)}\end{tabular} &\begin{tabular}[c]{@{}c@{}}\textbf{71.24}\\ \textbf{(+4.06)}\end{tabular} &  \begin{tabular}[c]{@{}c@{}} \textbf{72.32}\\ \textbf{(+4.51)}\end{tabular} & 4      \\ \bottomrule
\end{tabular}
}
\subcaption{CNN-based SNN on ImageNet dataset.\label{tab:imagenet-cnn}}
\end{subtable}
\begin{subtable}[c]{0.52\textwidth}
\centering
\resizebox{0.99\textwidth}{!}{\begin{tabular}{clcccc}
\toprule
\multicolumn{2}{c}{\multirow{3}{*}{\diagbox[dir=SE,width=3cm,height=0.8cm]{Method}{Architecture}}} & \multirow{2}{*}{\begin{tabular}[c]{@{}c@{}}Sformer-8-384\\ (\textit{P}: 16.81M)\end{tabular}} & \multirow{2}{*}{\begin{tabular}[c]{@{}c@{}}Sformer-8-512\\ (\textit{P}: 29.68M)\end{tabular}} & \multirow{2}{*}{\begin{tabular}[c]{@{}c@{}}Sformer-8-768\\ (\textit{P}: 66.34M)\end{tabular}} & \multirow{3}{*}{\begin{tabular}[c]{@{}c@{}}Time\\ Step\end{tabular}} \\
\multicolumn{2}{c}{}              &     &     &     &   \\
\multicolumn{2}{c}{}              & Top1 Acc.              & Top1 Acc.              & Top1 Acc.              &   \\ \midrule
\multirow{4}{*}{\rotatebox[origin=c]{90}{\parbox[c]{1.5cm}{\centering w/o \\ \textit{KD}}}}    & Spikformer~\cite{zhou2022spikformer}     & 70.24 & 73.38 & 74.81 & 4 \\
 & +CML~\cite{zhou2023enhancing} & 72.73 & 75.61 & 77.34 & 4 \\
 & Sformer~\cite{zhou2023spikingformer}        & 72.45 & 74.79 & 75.85 & 4 \\
 & +CML~\cite{zhou2023enhancing}    & 74.35 & \baseline{76.54} & 77.64 & 4 \\ \midrule
 & Teacher ANN & 81.78 & 81.78 & 81.78 & 1 \\ \midrule
\multirow{4}{*}{\rotatebox[origin=c]{90}{\parbox[c]{1cm}{\centering with \textit{KD}}}} & KDSNN~\cite{xu2023constructing}          &   \baseline{74.62}  & 76.44   &  \baseline{77.83}   & 4 \\
 & LaSNN~\cite{hong2023lasnn}          &  73.85   &   76.38  &  77.66   & 4 \\
 & \begin{tabular}[l]{@{}c@{}}BKDSNN\\ (ours)\end{tabular}  & \begin{tabular}[c]{@{}c@{}}\textbf{75.48}\\ \textbf{(+0.86)}\end{tabular} & \begin{tabular}[c]{@{}c@{}}\textbf{77.24}\\ \textbf{(+0.70)}\end{tabular} & \begin{tabular}[c]{@{}c@{}}\textbf{79.93}\\ \textbf{(+2.10)}\end{tabular} & 4 \\ \bottomrule
\end{tabular}}
\subcaption{Transformer-based SNN on ImageNet dataset.\label{tab:imagenet-transformer}}
\end{subtable}
\label{tab:imagenet}
\vskip -1cm
\end{table*}

\begin{table*}[t]
\caption{\textbf{(a) Comparison with Meta-SpikeFormer~\cite{yao2023spike} on ImageNet. (b) Experimental results on CIFAR10-DVS.} Mformer represents Meta-SpikeFormer~\cite{yao2023spike}. $\star$: SNN with high inference time-step~(\ie, 16) serves as the teacher model for corresponding SNN with low inference time-step~(\ie, 4 and 8). The prior state-of-the-art results are marked in \colorbox{baselinecolor}{gray}. Best results are in \textbf{bold}.}
\vskip -0.3cm
\begin{subtable}[c]{0.45\textwidth}
\centering
\resizebox{0.99\textwidth}{!}{\begin{tabular}{clcccc}
\toprule
\multicolumn{2}{c}{\multirow{3}{*}{\diagbox[dir=SE,width=3cm,height=0.8cm]{Method}{Architecture}}} & \multirow{2}{*}{\begin{tabular}[c]{@{}c@{}}Mformer-8-256\\ (\textit{P}: 15.11M)\end{tabular}} & \multirow{2}{*}{\begin{tabular}[c]{@{}c@{}}Mformer-8-384\\ (\textit{P}: 31.33M)\end{tabular}} & \multirow{2}{*}{\begin{tabular}[c]{@{}c@{}}Mformer-8-512\\ (\textit{P}: 55.35M)\end{tabular}} & \multirow{3}{*}{\begin{tabular}[c]{@{}c@{}}Time\\ Step\end{tabular}} \\
\multicolumn{2}{c}{}              &     &     &     &   \\
\multicolumn{2}{c}{}              & Top1 Acc.              & Top1 Acc.              & Top1 Acc.              &   \\ \midrule
\multirow{2}{*}{\rotatebox[origin=c]{90}{\parbox[c]{0.5cm}{\centering w/o \\ \textit{KD}}}}    & \multirow{2}{*}{Mformer~\cite{yao2023spike}}    & 71.82 & 75.40 & 77.96 & 1 \\
 & & 74.11 & 77.20 & 78.81 & 4 \\ \midrule
 & Teacher ANN & 81.78 & 81.78 & 81.78 & 1 \\ \midrule
\multirow{5}{*}{\rotatebox[origin=c]{90}{\parbox[c]{1cm}{\centering with \textit{KD}}}} & \multirow{2}{*}{Mformer~\cite{yao2023spike}}         &   --  & --   &  79.11   & 1 \\
 & &  --  &   --  &  \baseline{80.00}   & 4 \\
 & \multirow{3}{*}{\begin{tabular}[l]{@{}c@{}}BKDSNN\\ (ours)\end{tabular}}  & 72.93 & 76.10 & 79.82 & 1 \\ 
 & & \begin{tabular}[c]{@{}c@{}}\textbf{75.32}\\ \textbf{(+1.21)}\end{tabular} & \begin{tabular}[c]{@{}c@{}}\textbf{77.92}\\ \textbf{(+0.72)}\end{tabular} & \begin{tabular}[c]{@{}c@{}}\textbf{80.93}\\ \textbf{(+0.93)}\end{tabular} & 4 \\ \bottomrule
\end{tabular}}
\subcaption{Meta-SpikeFormer on ImageNet dataset.\label{tab:imagenet-mformer}}
\end{subtable}
\begin{subtable}[c]{0.55\textwidth}
\centering
\resizebox{0.99\textwidth}{!}{
\begin{tabular}{lcccc|lcccc}
\toprule
\multicolumn{5}{c}{Transformer-based}     & \multicolumn{5}{c}{CNN-based}  \\ 
Method       & Acc  & \begin{tabular}[c]{@{}c@{}}Time\\ Step\end{tabular} & Acc  & \begin{tabular}[c]{@{}c@{}}Time\\ Step\end{tabular} & Method      & Acc  & \begin{tabular}[c]{@{}c@{}}Time\\ Step\end{tabular} & Acc  & \begin{tabular}[c]{@{}c@{}}Time\\ Step\end{tabular} \\ \midrule
\begin{tabular}[c]{@{}c@{}}Sformer-2-256\\ +CML~\cite{zhou2023enhancing}$\star$\end{tabular} & 81.4 & 16  & 81.4 & 16  & \begin{tabular}[c]{@{}c@{}}Wide-7B-Net\\ \cite{fang2021deep}$\star$\end{tabular} & 74.4 & 16  & 74.4 & 16  \\ \midrule
\begin{tabular}[c]{@{}c@{}}Sformer-2-256\\ +CML~\cite{zhou2023enhancing}\end{tabular} &  76.4    & 4   &   79.2   & 8   & \begin{tabular}[c]{@{}c@{}}Wide-7B-Net\\ \cite{fang2021deep}\end{tabular} & 64.8 & 4   & 70.2 & 8   \\
KDSNN~\cite{xu2023constructing} &  78.1    & 4   &   79.9   & 8   & KDSNN~\cite{xu2023constructing}       &   66.1   & 4   &  \baseline{70.9}    & 8   \\
LaSNN~\cite{hong2023lasnn} &  \baseline{78.3}    & 4   &   \baseline{80.1}   & 8   & LaSNN~\cite{hong2023lasnn}       &   \baseline{66.6}  & 4   &   70.5   & 8   \\
\begin{tabular}[l]{@{}c@{}}BKDSNN\\ (ours)\end{tabular}       &  \textbf{\begin{tabular}[c]{@{}c@{}}79.3\\ (+1.0)\end{tabular}}    & 4   &  \textbf{\begin{tabular}[c]{@{}c@{}}80.8\\ (+0.7)\end{tabular}}    & 8   & \begin{tabular}[l]{@{}c@{}}BKDSNN\\ (ours)\end{tabular} & \textbf{\begin{tabular}[c]{@{}c@{}}68.3\\ (+1.7)\end{tabular}} & 4   & \textbf{\begin{tabular}[c]{@{}c@{}}72.2\\ (+1.3)\end{tabular}} & 8   \\ \bottomrule
\end{tabular}
}
\subcaption{Experimental results on CIFAR10-DVS.\label{tab:cifardvs}}
\end{subtable}
\vskip -0.75cm
\end{table*}

\begin{figure}[t]
    \centering
    \subcaptionbox{Feature map visualization.\label{fig:fmap}}
   {\includegraphics[width=0.64\linewidth]{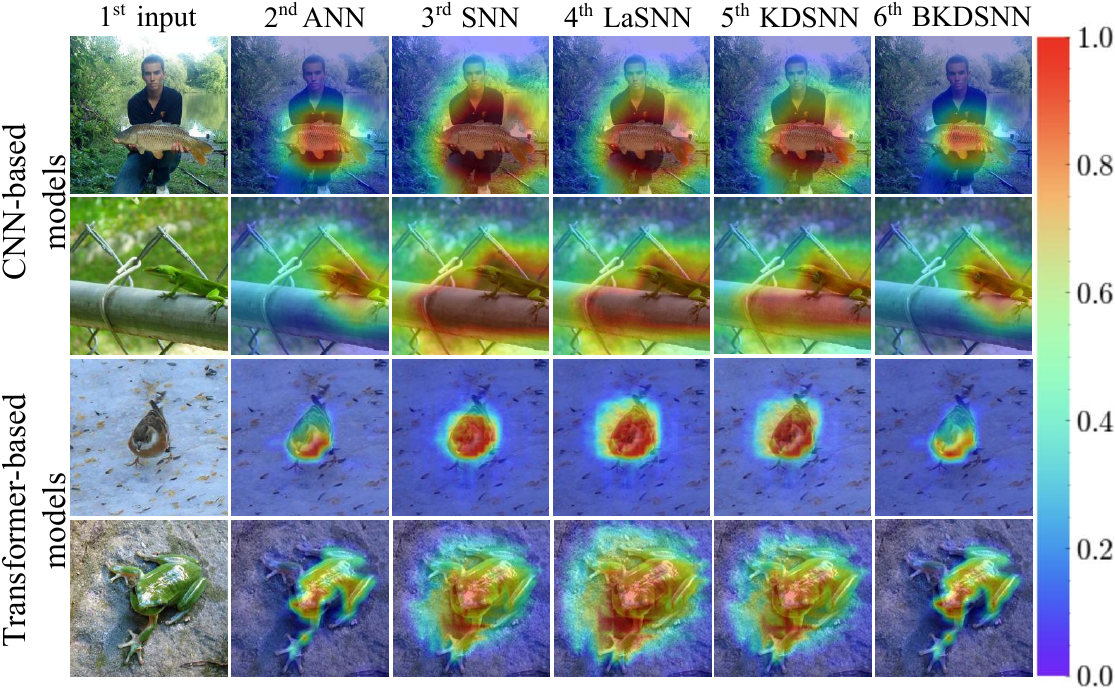}}
   \subcaptionbox{GPU hours on ImageNet.\label{fig:gpuhours}}
   {\includegraphics[width=0.35\linewidth]{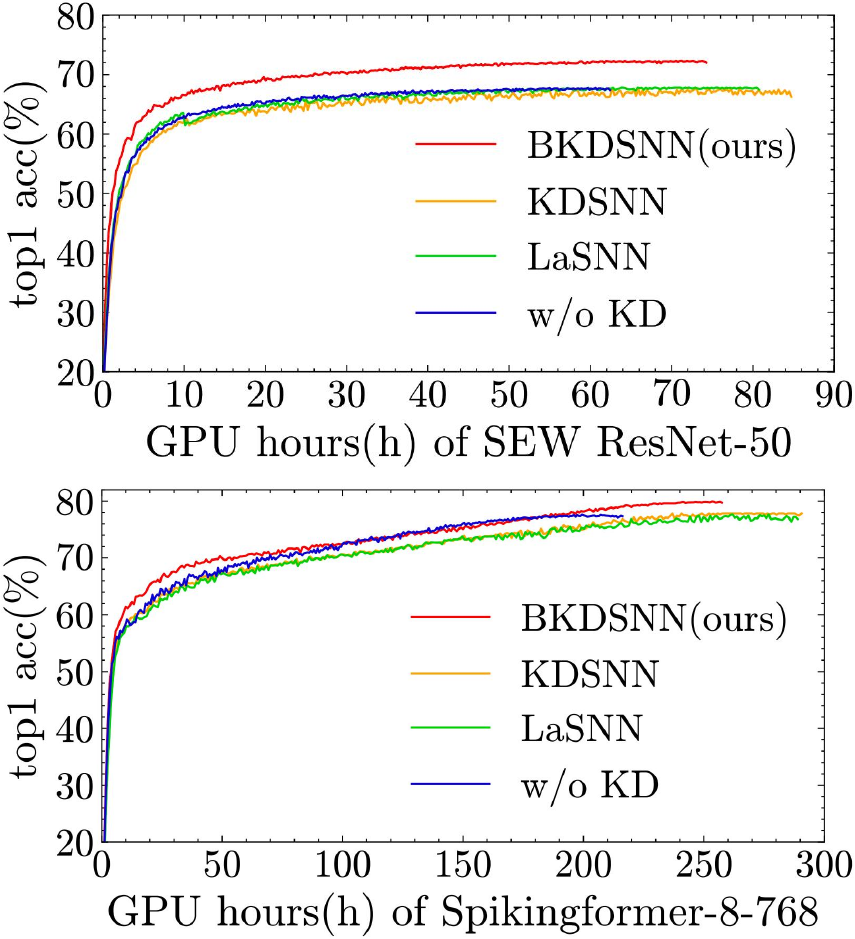}}
\vskip -0.3cm
    \caption{(a) \textbf{Feature map visualization} of different methods on SEW ResNet-18 (CNN) and Sformer-8-384+CML (Transformer). Input images are sampled from the ImageNet validation dataset. Shaded colors from blue to red indicate the impacts of the regions on the classification scores from low to high. (b) \textbf{Top1 accuracy versus GPU hours on ImageNet} with SEW ResNet-50 and Spikingformer-8-768.}
    \vskip -0.7cm
\end{figure}



\paragraph{Comparison on ImageNet} between BKDSNN and previous methods are tabulated in
\cref{tab:imagenet-cnn}, \cref{tab:imagenet-transformer} and~\cref{tab:imagenet-mformer}, 
As anticipated, BKDSNN outperforms previous SOTA methods with both CNN- and Transformer-based models on ImageNet. 
BKDSNN achieves 72.32\% top1 accuracy with SEW ResNet-50 and 80.93\% with Mformer-8-512, with 4.51\% and 0.93\% improvement, respectively.

As reported in~\cref{tab:imagenet-cnn}, \cref{tab:imagenet-transformer} and~\cref{tab:imagenet-mformer}, ANN outperforms previous learning-based SNN by a large margin, which will serve as the teacher model to provide adequate knowledge for the student SNN. 
Based on our experiments, the accuracy improvement of SNN trained by KDSNN~\cite{xu2023constructing} and LaSNN~\cite{hong2023lasnn} is limited, due to the inappropriate distillation between student and teacher models. 
Nevertheless, thanks to the blurred student features that effectively mimic and restore the teacher features, BKDSNN achieves promising results on the large-scale dataset (\ie ImageNet).

\paragraph{Feature Visualization.} 
To further interpret BKDSNN, we apply Grad-CAM~\cite{Selvaraju_2019} to visualize the intermediate feature before the last layer using different methods. In $\text{2}^\text{nd}$~(ANN) to $\text{3}^\text{rd}$~(SNN) column of \cref{fig:fmap}, ANN properly focuses on the regions with vital information (highlighted in red), while learning-based SNN is distracted by the background area. 
Such erroneous attention of SNN ($\text{3}^\text{rd}$ column in \cref{fig:fmap}) does not get calibrated by the previous KD methods, \ie, KDSNN~\cite{xu2023constructing} ($\text{4}^\text{th}$ column) and LaSNN~\cite{hong2023lasnn} ($\text{5}^\text{th}$ column).
However, the attention (shaded) region of our BKDSNN~($\text{6}^\text{th}$ column) is quite similar to the pre-trained ANN, which demonstrates the effectiveness of BKD in learning the knowledge of ANN by mimicking and restoring teacher ANN features with blurred student SNN features.

\paragraph{Comparison of Training Cost.}
We plot the evolution curves of validation top1 accuracy during GPU hours as depicted in~\cref{fig:gpuhours}. Without applying knowledge distillation~(denoted as w/o KD in blue lines), the training of SEW ResNet-50 and Spikingformer-8-768 consume the least GPU hours. With the introduction of layer-wise distillation~(LaSNN~\cite{hong2023lasnn} in green lines and KDSNN~\cite{xu2023constructing} in yellow lines), the training of SEW ResNet-50 and Spikingformer-8-768 consume more hours while the improvement is limited. Compared with the layer-wise distillation methods above, our single-layer BKDSNN~(in red lines) saves GPU hours and boosts the performance significantly.

\subsection{Comparison on Neuromorphic Dataset}

We also evaluate BKDSNN on CIFAR10-DVS with results reported in~\cref{tab:cifardvs}. Specifically, BKDSNN outperforms previous SOTA results by 1.0\% on models with 4 time-steps and 0.7\% on models with 8 time-steps. Note that, the improvement of BKDSNN on Wide-7B-Net~\cite{fang2021deep} is more pronounced than that on Sformer-2-256~\cite{zhou2023enhancing}, which is probably due to the fact that Wide-7B-Net~\cite{fang2021deep} with 16 inference time-steps provides more knowledge for the corresponding student model with lower time-steps, compared to Sformer-2-256~\cite{zhou2023enhancing}.

\subsection{Ablation Study}
\paragraph{Ablation on multiple losses~($\mathcal{L}_\textrm{LD}$, $\mathcal{L}_\textrm{BKD}$ and $\mathcal{L}_\textrm{MD}$).} 
We test different settings of BKDSNN, including versions without any knowledge distillation (w/o KD), with logit-based KD ($+\mathcal{L}_\textrm{LD}$), with BKD ($+\mathcal{L}_\textrm{BKD}$), with mixed distillation (logit-based + BKD, $+\mathcal{L}_\textrm{MD}$). Results are tabulated in \cref{tab:ablation-distillation}.
Compared to logit-based KD, applying BKD to the input feature of the last layer shows on-par accuracy improvement on both CNN- and Transformer-based SNNs.
Furthermore, thanks to the compatibility between the blurred feature-based KD and logit-based KD, their mixed version shows the maximal accuracy improvement.  


\begin{table*}[t]
\caption{
\textbf{Ablation study with multiple settings}. \textit{R}, \textit{S}, and \textit{KD} denotes ResNet, Sformer and Knowledge Distillation respectively. \textit{S}-4-384-400E refers to Sformer-4-384 after 400 epochs training. $\star$: results from~\cite{fang2021deep}, $\diamond$: results from~\cite{zhou2023enhancing}. Best results are in \textbf{bold}, runner-up results are marked in \colorbox{baselinecolor}{gray}.
}
\begin{subtable}[c]{0.46\textwidth}
\centering
\resizebox{0.99\textwidth}{!}{
\begin{tabular}{lcccccc}
\toprule
\multirow{2}{*}{Method} & \multicolumn{2}{c}{Config}    & \multicolumn{2}{c}{\textit{S}-4-384-400E}        & \multicolumn{2}{c}{\textit{R}-19}  \\
      & $\tau$     & $\lambda$  & \multicolumn{1}{c}{\begin{tabular}[c]{@{}c@{}}\textit{CF}10\\ Acc\end{tabular}}              & \multicolumn{1}{c}{\begin{tabular}[c]{@{}c@{}}\textit{CF}100\\ Acc\end{tabular}}               & \multicolumn{1}{c}{\begin{tabular}[c]{@{}c@{}}\textit{CF}10\\ Acc\end{tabular}}               & \multicolumn{1}{c}{\begin{tabular}[c]{@{}c@{}}\textit{CF}100\\ Acc\end{tabular}}              \\ \midrule
\multirow{3}{*}{$+\mathcal{L}_\text{LD}$}       & 1.5   & -      & 96.13  & 81.40  & 94.50  & 74.66  \\
      & 2.0   & -      & 96.17  & 81.25  & 94.53  & 74.57  \\
      & 4.0   & -      & 96.31  & 80.88  & 94.56  & 74.52  \\ \midrule
\multirow{3}{*}{$+\mathcal{L}_\text{BKD}$}      & -     & 0.15   & 96.13  & 81.38  & 94.52  & 74.76  \\
      & -     & 0.40   & 96.26  & 81.43  & 94.56  & 74.83  \\
      & -     & 0.60   & 96.31  & 81.53  & 94.61  & 74.89  \\ \midrule
\multirow{4}{*}{$+\mathcal{L}_\text{MD}$}              & 2.0   & 0.15   & 96.18  & 81.63  & 94.64  & 74.95  \\
      & 2.0   & 0.40   & \baseline{96.20}  & \baseline{81.67}  & \baseline{94.68}  & \baseline{74.98}  \\
      & \multirow{2}{*}{2.0} & \multirow{2}{*}{0.60} & \multirow{2}{*}{\textbf{\begin{tabular}[c]{@{}c@{}}96.37\\ (+0.11)\end{tabular}}} & \multirow{2}{*}{\textbf{\begin{tabular}[c]{@{}c@{}}81.86\\ (+0.19)\end{tabular}}} & \multirow{2}{*}{\textbf{\begin{tabular}[c]{@{}c@{}}94.76\\ (+0.08)\end{tabular}}} & \multirow{2}{*}{\textbf{\begin{tabular}[c]{@{}c@{}}75.07\\ (+0.09)\end{tabular}}} \\
      &       &        &        &        &        &        \\ \bottomrule
\end{tabular}
}
\subcaption{Ablation of $\tau$ and $\lambda$ on CIFAR. \label{tab:ablation-hyperparameter}}
\end{subtable}
\begin{subtable}[c]{0.54\textwidth}
\centering
\resizebox{0.97\textwidth}{!}{
\begin{tabular}{lcccccc}
\toprule
\multirow{3}{*}{Method}          & \multicolumn{3}{c}{CNN-based}       & \multicolumn{3}{c}{Transformer-based}   \\
      & {\begin{tabular}[c]{@{}c@{}} Student\\ SNN\end{tabular}} & \begin{tabular}[c]{@{}c@{}} Teacher\\ Acc\end{tabular} & {\begin{tabular}[c]{@{}c@{}} SNN\\ Acc\end{tabular}} & {\begin{tabular}[c]{@{}c@{}} Student\\ SNN\end{tabular}} & {\begin{tabular}[c]{@{}c@{}} Teacher\\ Acc\end{tabular}} & {\begin{tabular}[c]{@{}c@{}} SNN\\ Acc\end{tabular}}\\ \midrule
\multirow{3}{*}{\begin{tabular}[c]{@{}c@{}} w/o\\ \textit{KD}\end{tabular}} & \textit{R}-18$\star$              & -              & 63.18 & \textit{S}-8-384$\diamond$  & -  & 74.35    \\
      & \textit{R}-34$\star$ & -              & 67.04 & \textit{S}-8-512$\diamond$ & -  & 76.54   \\
      & \textit{R}-50$\star$ & -              & 67.78 & \textit{S}-8-768$\diamond$ & -  & 77.64     \\ \midrule
\multirow{3}{*}{$+\mathcal{L}_\text{LD}$}  & \textit{R}-18      & 69.76          & \baseline{65.36(+2.18)} & \textit{S}-8-384 & \multirow{3}{*}{81.78} & 74.69(+0.24)  \\
      & \textit{R}-34 & 73.31          & 69.90(+2.86) & \textit{S}-8-512          &    & 76.99(+0.45)    \\
      & \textit{R}-50      & 76.13          & \baseline{71.78(+4.00)} & \textit{S}-8-768          & & 79.04(+1.40)   \\ \midrule
\multirow{3}{*}{$+\mathcal{L}_\text{BKD}$}    & \textit{R}-18 & 69.76          & 65.16(+1.98) & \textit{S}-8-384 & \multirow{3}{*}{81.78} & \baseline{75.08(+0.73)}    \\
      & \textit{R}-34 & 73.31          & \baseline{70.23(+3.19)} & \textit{S}-8-512 &    & \textbf{77.29(+0.75)}   \\
      & \textit{R}-50 & 76.13          & 71.24(+3.46) & \textit{S}-8-768 &    & \baseline{79.64(+2.00)}\\ \midrule
\multirow{3}{*}{$+\mathcal{L}_\text{MD}$} & \textit{R}-18 & 69.76          & \textbf{65.60(+2.42)}   & \textit{S}-8-384 & \multirow{3}{*}{81.78} & \textbf{75.48(+1.13)}      \\
       & \textit{R}-34 & 73.31          & \textbf{71.24(+4.20)} & \textit{S}-8-512 &    & \baseline{77.24(+0.70)}\\
       & \textit{R}-50 & 76.13          &\textbf{72.32(+4.54)} & \textit{S}-8-768 &    & \textbf{79.93(+2.29)}  \\ \bottomrule
\end{tabular}
}
\subcaption{Ablation of multiple losses on ImageNet.\label{tab:ablation-distillation}}
\end{subtable}
\vskip -0.2cm

\begin{subtable}[c]{0.37\textwidth}
\centering
\resizebox{\textwidth}{!}{
\begin{tabular}{lccc}
\toprule
\multicolumn{1}{l}{\multirow{2}{*}{Method}}                & \multirow{2}{*}{\begin{tabular}[c]{@{}c@{}}SEW \textit{R}-18\\ Top1 Acc\end{tabular}} & \multirow{2}{*}{\begin{tabular}[c]{@{}c@{}}\textit{S}-8-384\\ Top1 Acc\end{tabular}} & \multirow{2}{*}{\begin{tabular}[c]{@{}c@{}}\textit{KD}\\ location\end{tabular}} \\
\multicolumn{1}{c}{}   &                 &                &  \\ \midrule
LaSNN~\cite{hong2023lasnn} & 63.31           & 73.85          & layer-wise \\
+$\mathcal{L}_\text{BKD}$ & 63.82  & 74.47 & layer-wise \\ \hline
KDSNN~\cite{xu2023constructing} & 63.42           & 74.62          & layer-wise \\
+$\mathcal{L}_\text{BKD}$ & \baseline{63.94}  & 74.93 & layer-wise \\ \midrule
KDSNN~\cite{xu2023constructing} & 63.87           & \baseline{74.96}          & single-layer \\
\begin{tabular}[l]{@{}c@{}}BKDSNN\\ (ours)\end{tabular}                 & \textbf{\begin{tabular}[c]{@{}c@{}}65.60\\ (+1.66)\end{tabular}}  & \textbf{\begin{tabular}[c]{@{}c@{}}75.48\\ (+0.52)\end{tabular}} & single-layer \\ \bottomrule
\end{tabular}
}
\subcaption{Ablation of BKD on ImageNet.\label{tab:ablation_kd_loc}}
\end{subtable}
\begin{subtable}[c]{0.63\textwidth}
\centering
\resizebox{0.98\textwidth}{!}{
\begin{tabular}{cccccccccc}
\toprule
\multirow{3}{*}{Method} & \multirow{3}{*}{$\lambda$} & \multicolumn{4}{c}{Spikingformer-4-384-400E}   & \multicolumn{4}{c}{ResNet-19} \\& & \multicolumn{2}{c}{\begin{tabular}[c]{@{}c@{}}\textit{CIFAR}10\\ Acc\end{tabular}} & \multicolumn{2}{c}{\begin{tabular}[c]{@{}c@{}}\textit{CIFAR}100\\ Acc\end{tabular}} & \multicolumn{2}{c}{\begin{tabular}[c]{@{}c@{}}\textit{CIFAR}10\\ Acc\end{tabular}} & \multicolumn{2}{c}{\begin{tabular}[c]{@{}c@{}}\textit{CIFAR}100\\ Acc\end{tabular}} \\& & w $\mathcal{G}$    & w/o $\mathcal{G}$ & w $\mathcal{G}$ & w/o $\mathcal{G}$ & w $\mathcal{G}$    & w/o $\mathcal{G}$ & w $\mathcal{G}$    & w/o $\mathcal{G}$  \\ \midrule
\multirow{4}{*}{$+\mathcal{L}_\text{BKD}$}   & 0.00 & 95.38 & 94.38 & 80.44  & 76.64 & 94.47 & 93.42 & 74.68 & 70.97 \\& 0.15 & 96.13  & 94.60 & 81.38   & 76.81 & 94.52  & 93.58 & 74.76  & 71.22 \\& 0.40 & \cellcolor{baselinecolor}96.26  & \cellcolor{baselinecolor}94.82 & \cellcolor{baselinecolor}81.43   & \cellcolor{baselinecolor}77.14 & \cellcolor{baselinecolor}94.56  & \cellcolor{baselinecolor}93.64 & \cellcolor{baselinecolor}74.83  & \cellcolor{baselinecolor}71.29 \\& 0.60 & \textbf{96.31}  & \textbf{94.89} & \textbf{81.53}   & \textbf{77.16} & \textbf{94.61}  & \textbf{93.72} & \textbf{74.89}  & \textbf{71.37}\\ \bottomrule
 
\end{tabular}
}
\subcaption{Ablation of $\lambda$ and $\mathcal{G}$ on CIFAR. \label{tab:ablation-cifar-G}}
\end{subtable}
\label{tab:ablation}
\vskip -1cm
\end{table*}

\begin{figure*}[t]
\centering
\subcaptionbox{SEW ResNet-18}{\includegraphics[width=4cm]{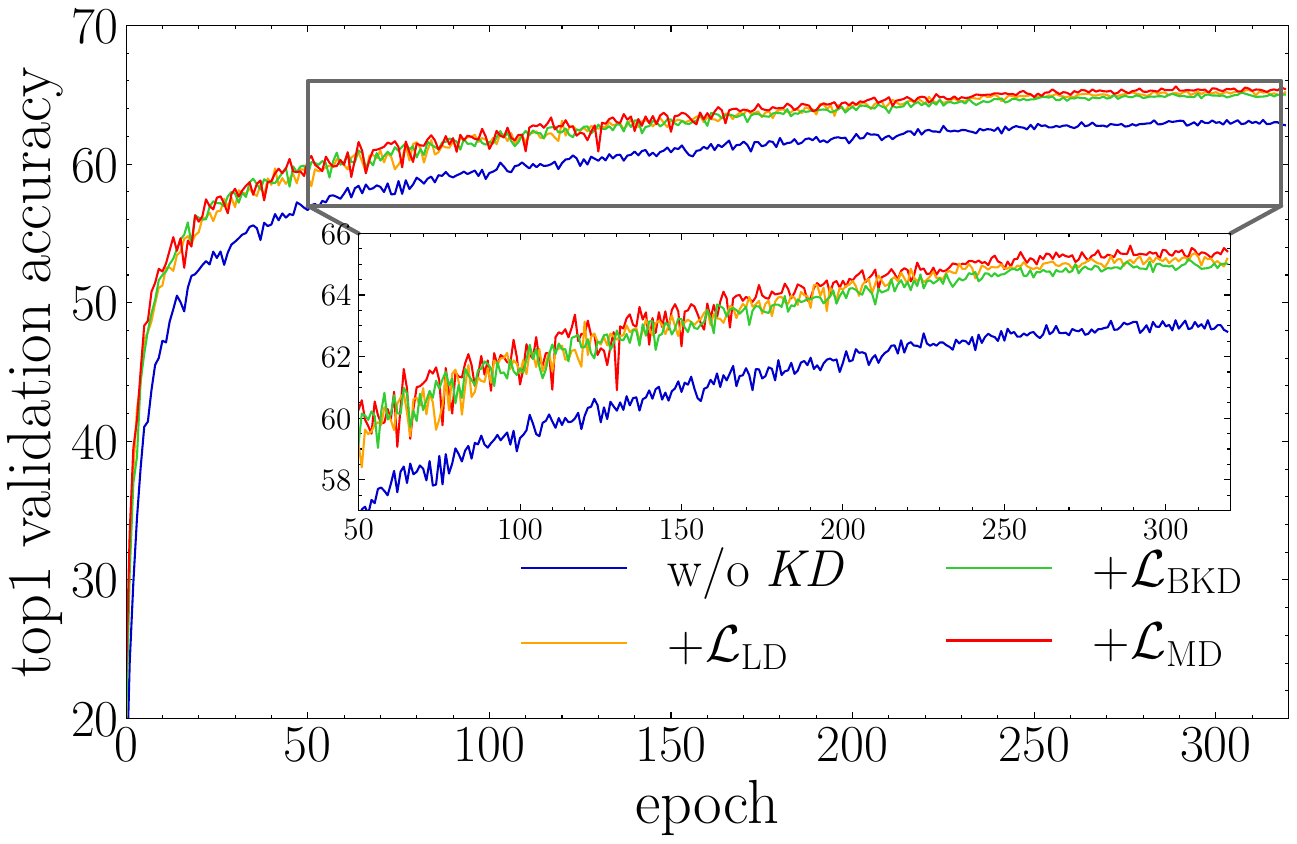}}
\subcaptionbox{SEW ResNet-34}{\includegraphics[width=4cm]{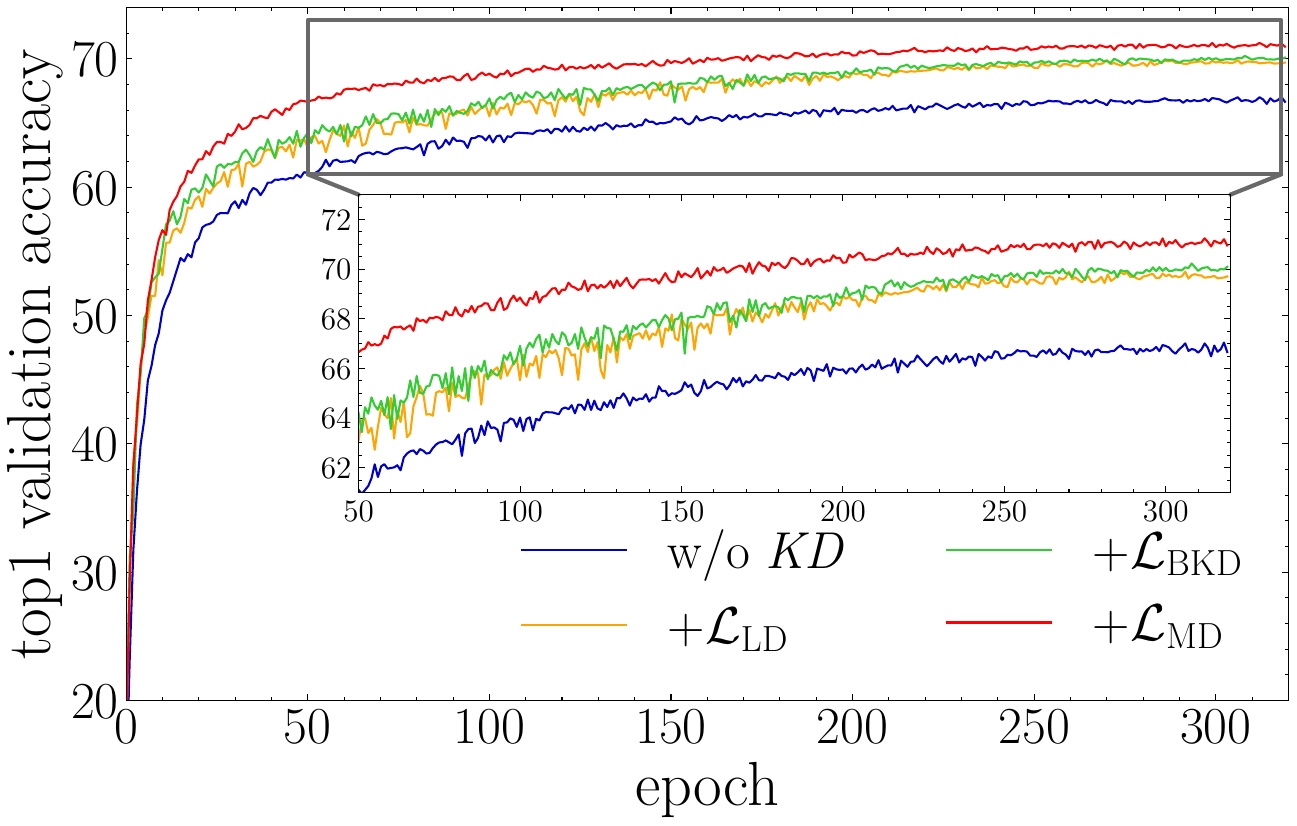}}
\subcaptionbox{SEW ResNet-50}{\includegraphics[width=4cm]{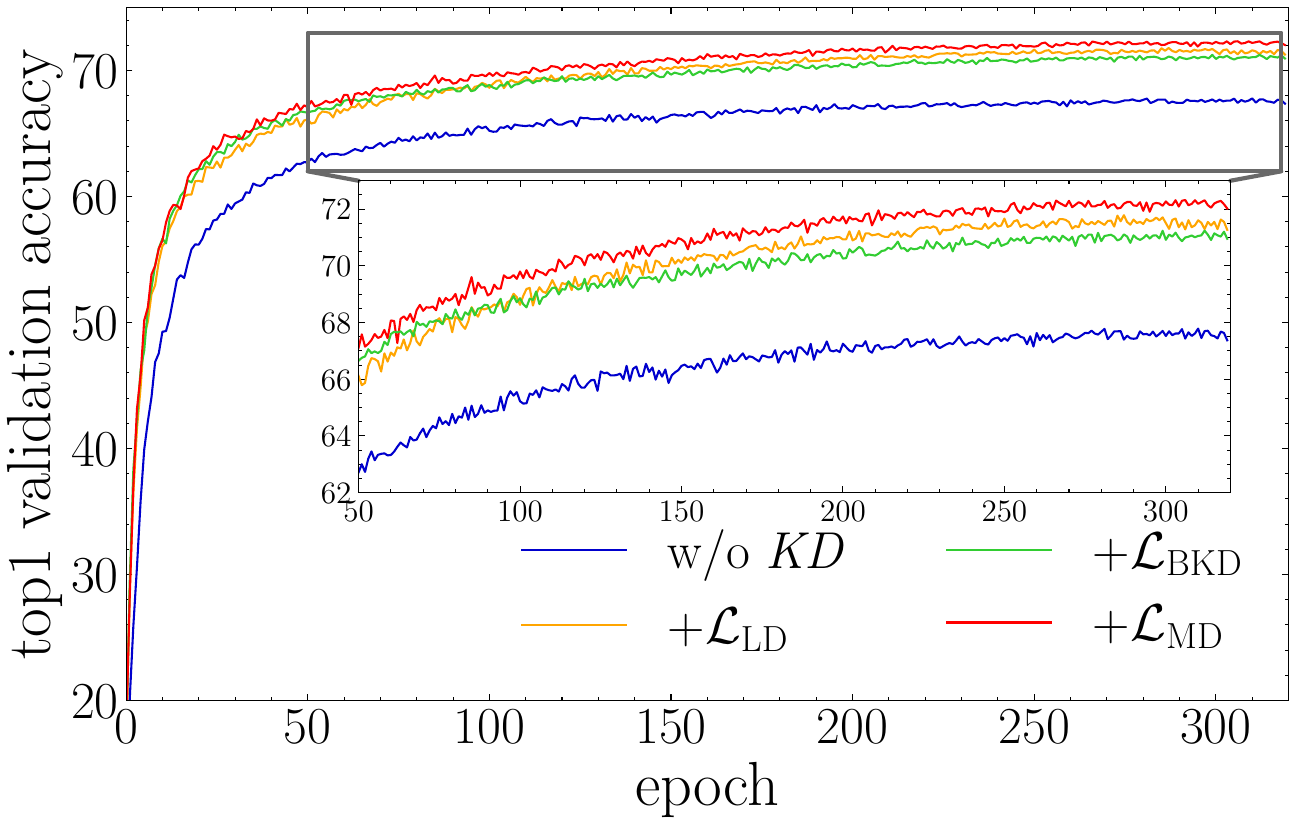}} \\
\subcaptionbox{Sformer-8-384}{\includegraphics[width=4cm]{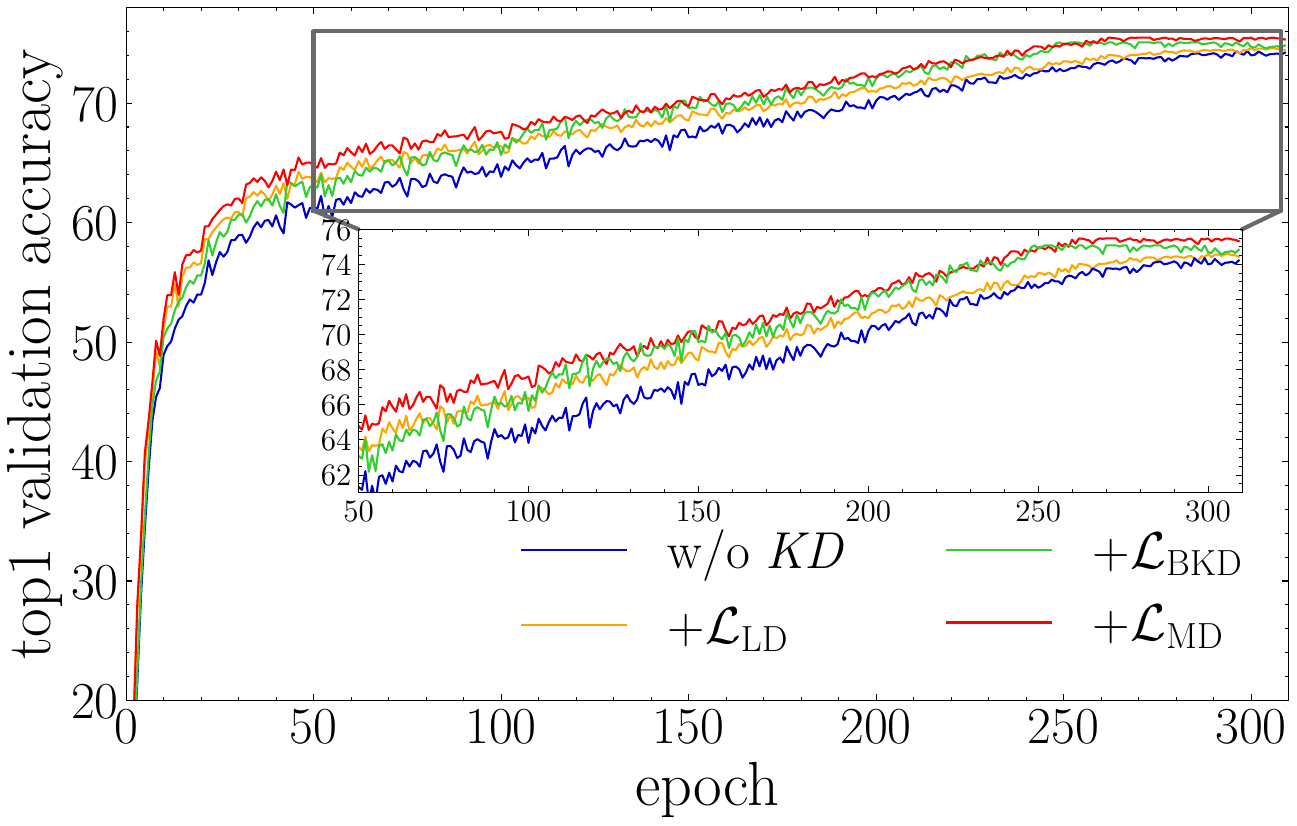}}
\subcaptionbox{Sformer-8-512}{\includegraphics[width=4cm]{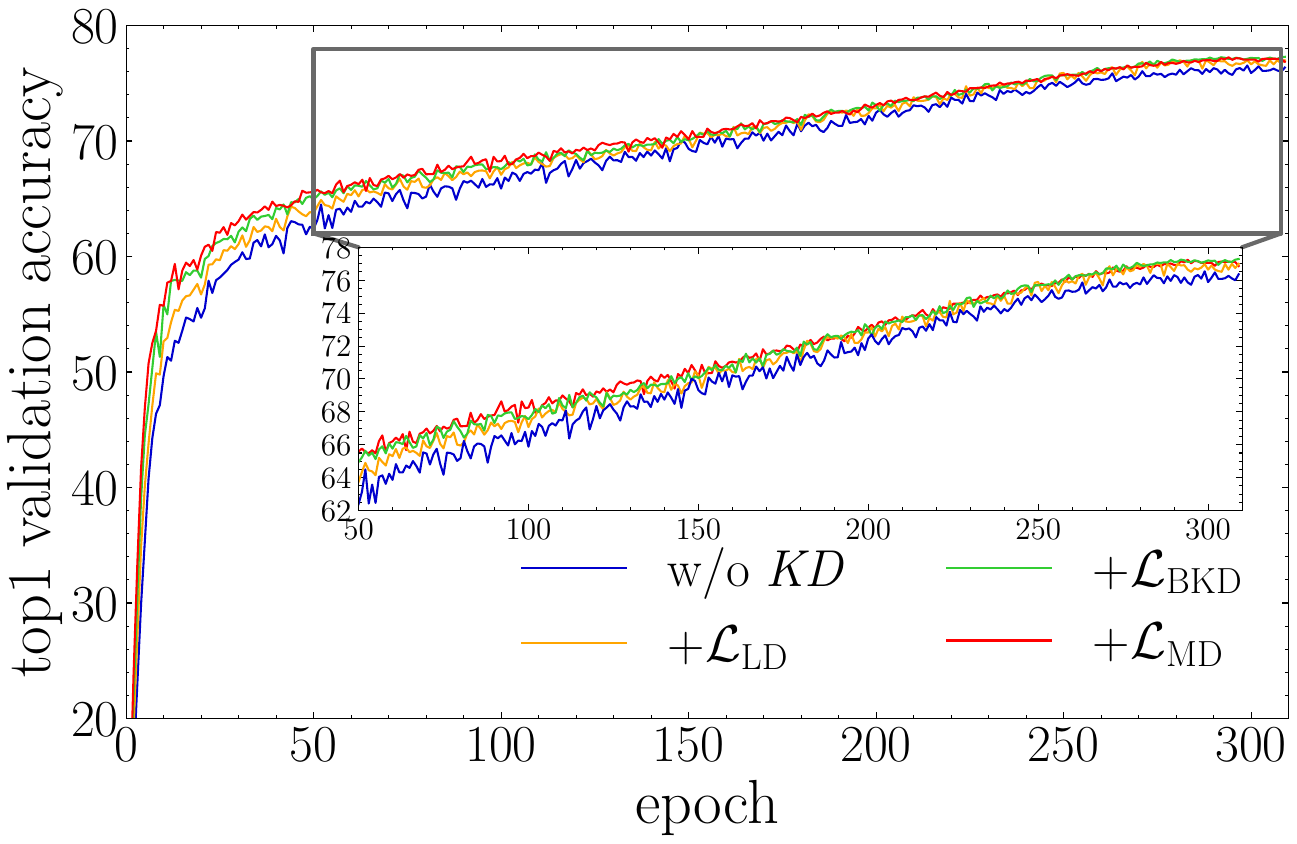}}
\subcaptionbox{Sformer-8-768}{\includegraphics[width=4cm]{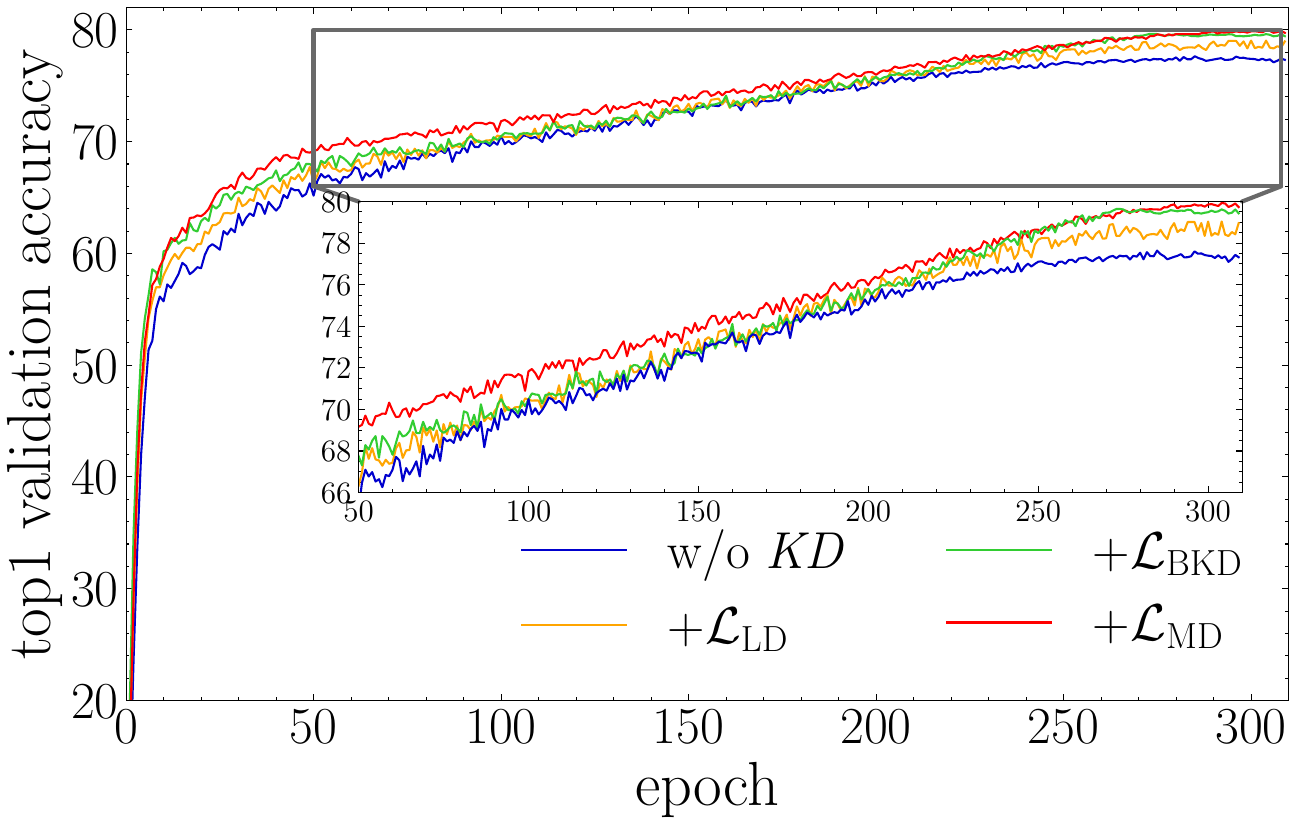}} 
 
\vskip -0.2cm
\caption{\textbf{Evolution curve of validation accuracy during training}, with CNN- (top 3 sub-figures) and Transformer-based (bottom 3 sub-figures) SNNs on ImageNet. The labels in the legends are the same as in \cref{tab:ablation-distillation}.}
\label{fig:training_curve}
\vskip -0.8cm
\end{figure*}

In addition to the accuracy obtained for SEW ResNet-18 and Sformer-8-384 discussed above, we also plot their evolution curves of validation accuracy during training as in \cref{fig:training_curve}.
We observe that with the introduction of logits-based distillation $\mathcal{L}_\text{LD}$~(yellow curve) and feature-based blurred knowledge distillation $\mathcal{L}_\text{BKD}$~(green curve), the performance is significantly improved compared to the vanilla counterpart~(blue curve).
Moreover, as anticipated, the mixed distillation (red curve) in KDSNN leads to faster convergence with higher validation accuracy during the training.
These experimental results highlight the important role of logits-based and feature-based knowledge from the teacher ANN in improving the performance of student SNNs.

\paragraph{Ablation of BKD and its location}
We have reported the KDSNN accuracy under different settings in \cref{tab:ablation-distillation} where KD was applied to logits, the last layer feature, or both.
Two more questions still remain unanswered: 1) whether BKD outperforms the vanilla implementation? 2) Is applying KD/BKD to the last layer better than the layer-wise implementation?

To answer these questions, we perform experiments whose results are reported in \cref{tab:ablation_kd_loc}.
LaSNN~\cite{hong2023lasnn} and KDSNN~\cite{xu2023constructing} are two recent works of layer-wise implementation of KD for SNN training.
We replace the original feature-based distillation loss in LaSNN and KDSNN with BKD loss $\mathcal{L}_\text{BKD}$.
In LaSNN, the replacement improves accuracy by 0.51\% on SEW ResNet-18 and by 0.62\% on Sformer-8-384. 
In KDSNN, the replacement improves accuracy by 0.52\% on SEW ResNet-18 and by 0.31\% on Sformer-8-384. 
These experiments confirm the advantages of the blurred implementation of KD.
Moreover, we further test layer-wise (every layer feature) or single-layer (intermediate feature before the last layer) implementations of KD. We observe similar accuracy boosts for the two implementations. Thus, we conclude that applying BKD only to the intermediate feature map before the last layer is the adequate setup.

\paragraph{Ablation of temperature coefficient, blurred ratio and restoration block.} 
To examine the optimal setting of hyper-parameters, \ie, temperature $\tau$ in \cref{eq:flatten,eq:wsld}, blurred ration $\lambda$ in \cref{eq:bm} and restoration block $\mathcal{G}$ in~\cref{eq:restoration}, we sweep $\tau$, $\lambda$ and $\mathcal{G}$ with ResNet-19 and Sformer-4-384, under varying BKDSNN settings.
The results of the CIFAR-10 and CIFAR-100 datasets are reported in \cref{tab:ablation-hyperparameter} and~\cref{tab:ablation-cifar-G}.
Note that we increase the training epochs of Sformer-4-384 to 400 for more distinct results.
For logits-based distillation $\mathcal{L}_\text{LD}$, we find that flatter logits distribution~(with higher $\tau$) is better for classification with smaller categories~(CIFAR10), while sharper logits distribution~(with lower $\tau$) is more effective for classification with larger categories~(CIFAR100).
For BKD $\mathcal{L}_\text{BKD}$ and mixed distillation $\mathcal{L}_\text{MD}$, training with a higher blurred ratio $\lambda$ (\ie, 0.60) achieves greater accuracy across various datasets and models. As reported in~\cref{tab:ablation-cifar-G}, with the introduction of blurred ration $\lambda$ and restoration block $\mathcal{G}$, student SNN better mimics teacher ANN and conveys more sufficient information, which enhances performance on CIFAR-10 and CIFAR-100 with both CNN- and Transformer-based models.
\subsection{Firing Rate Analysis}

\paragraph{Firing Rate.} We compare the firing rate $r$ of the intermediate feature before last layer on CIFAR with Sformer-4-384 and ResNet-19 under multiple settings. The firing rate $r$ is defined as:
\begin{equation}
r = \frac{1}{bc^\text{stu}hw}\displaystyle\sum_{i}^{b}\sum_{j}^{c^\text{stu}}\sum_{k}^{h}\sum_{l}^{w} \Tilde{\tF}_{i, j, k, l}^\text{stu}
\end{equation}
\vskip -0.3cm
As reported in~\cref{tab:fr-cifar}, it reveals that on both CNN- and Transformer-based models, setting a higher firing rate $r$ achieves higher accuracy. Note that, with the introduction of $\mathcal{L}_\text{BKD}$ and $\mathcal{L}_\text{MD}$, high blurred ratio $\lambda$ encourages model to fire more spikes. 
Furthermore, compared with $\mathcal{L}_\text{LD}$, our $\mathcal{L}_\text{BKD}$ reduces the firing rate while improving the performance, which assists the energy consumption reduction. Meanwhile, in virtue of $\mathcal{L}_\text{MD}$, BKDSNN significantly improves the performance with a slight boost on firing rate, which leads to state-of-the-art. 

\subsection{Energy Consumption Analysis}

\begin{table*}[t]
\caption{Firing rate $r$ under multiple settings on CIFAR and Energy consumption of ANN and SNN on ImageNet. \textit{T} denotes Transformer.}
\vskip -0.3cm
\begin{subtable}[c]{0.45\textwidth}
\centering
\resizebox{0.99\textwidth}{!}{\begin{tabular}{lcccccccccc}
\toprule
\multirow{2}{*}{Method} & \multicolumn{2}{c}{Config} & \multicolumn{4}{c}{\textit{S}-4-384-400E}  & \multicolumn{4}{c}{\textit{R}-19}    \\
 &  $\tau$   &   $\lambda$   & \begin{tabular}[c]{@{}c@{}}\textit{CF}10\\ r~($e^{-2}$)\end{tabular} & \begin{tabular}[c]{@{}c@{}}\textit{CF}10\\ Acc\end{tabular} & \begin{tabular}[c]{@{}c@{}}\textit{CF}100\\ r~($e^{-2}$)\end{tabular} & \begin{tabular}[c]{@{}c@{}}\textit{CF}100\\ Acc\end{tabular} & \begin{tabular}[c]{@{}c@{}}\textit{CF}10\\ r~($e^{-2}$)\end{tabular} & \begin{tabular}[c]{@{}c@{}}\textit{CF}10\\ Acc\end{tabular} & \begin{tabular}[c]{@{}c@{}}\textit{CF}100\\ r~($e^{-2}$)\end{tabular} & \begin{tabular}[c]{@{}c@{}}\textit{CF}100\\ Acc\end{tabular} \\ \midrule
 \multirow{3}{*}{$+\mathcal{L}_\text{LD}$}      & 1.5   & - & 5.55      & 96.13 & \baseline{1.52} & 81.40  & 3.94      & 94.50 & 0.93 & 74.66  \\
 & 2.0   & - & 5.72      & 96.17 & 1.32 & 81.25  & 4.03      & 94.53 & 1.01 & 74.57  \\
 & 4.0   & - & 6.07      & \baseline{96.31} & 1.18 & 80.88  & \baseline{4.11}      & 94.56 & \baseline{1.08} & 74.52  \\ \midrule
\multirow{3}{*}{$+\mathcal{L}_\text{BKD}$}      & -   & 0.15 & 4.67      & 96.13 & 0.97 & 81.38  & 3.82      & 94.52 & 0.82 & 74.76  \\
 & -   & 0.40 & 5.30      & 96.26 & 1.03 & 81.43  & 3.94      & 94.56 & 0.89 & 74.83  \\
 & -   & 0.60 & 5.55      & \baseline{96.31} & 1.14 & 81.53  & 4.08      & 94.61 & 1.02 & 74.89  \\ \midrule
\multirow{3}{*}{$+\mathcal{L}_\text{MD}$}      & 2.0 & 0.15 & 5.39      & 96.18 & 1.14 & 81.63  & 3.92      & 94.64 & 1.03 & 74.95  \\
 & 2.0 & 0.40 & \baseline{6.17}      & 96.20 & 1.34 & \baseline{81.67}  & 3.99      & \baseline{94.68} & 1.07 & \baseline{74.98}  \\
 & 2.0 & 0.60 & \textbf{6.66}     & \textbf{96.37}      & \textbf{1.55}      & \textbf{81.86}       & \textbf{4.18}     & \textbf{94.76}      & \textbf{1.13}      & \textbf{75.07}       \\ \bottomrule
\end{tabular}}
\subcaption{Firing rate $r$ on CIFAR.\label{tab:fr-cifar}}
\end{subtable}
\begin{subtable}[c]{0.55\textwidth}
\centering
\resizebox{0.99\textwidth}{!}{
\begin{tabular}{lcccc}
\toprule
   Method      & \begin{tabular}[c]{@{}c@{}}Time\\Step\end{tabular} & \begin{tabular}[c]{@{}c@{}}SOPS(G)\\($\downarrow$)\end{tabular}        & \begin{tabular}[c]{@{}c@{}}Energy Consum-\\ ption~(mJ)($\downarrow$)\end{tabular} & \begin{tabular}[c]{@{}c@{}}Top1\\Acc~($\uparrow$)\end{tabular}       \\ \midrule
\textit{R}-50~\cite{he2016deep} & 1         & 4.12 (FLOPS)           & 18.95            & 76.13           \\
 Spiking \textit{R}-50~\cite{fang2021deep} & 4         & 5.31(28.88\%$\uparrow$)  & 5.44(71.29\%$\downarrow$)   & 57.66(-13.47) \\
 SEW \textit{R}-50~\cite{fang2021deep} & 4         & \baseline{4.83(17.23\%$\uparrow$)}  & \baseline{4.89(74.20\%$\downarrow$)}   & \baseline{67.78(-8.35)}  \\
 BKDSNN~(ours) & 4 & \textbf{4.81(16.75\%$\uparrow$)}  & \textbf{4.85(74.40\%$\downarrow$)}   & \textbf{72.32(-3.81)} \\ \midrule
\textit{T}-8-768~\cite{dosovitskiy2020image} & 1         & 18.60 (FLOPS)          & 85.56            & 81.16           \\
\textit{S}-8-768~\cite{zhou2023spikingformer} & 4         & \baseline{12.54(32.58\%$\downarrow$)} & \baseline{13.68(84.01\%$\downarrow$)}  & 75.85(-5.31)  \\
\textit{S}-8-768+CML~\cite{zhou2023enhancing} & 4         & \textbf{12.30(33.87\%$\downarrow$)} & \textbf{13.42(84.32\%$\downarrow$)}  & \baseline{77.64(-3.52)}  \\
 BKDSNN~(ours) & 4 & 12.60(32.26\%$\downarrow$) & 13.72(83.96\%$\downarrow$)  & \textbf{79.93(-1.23)} \\ \bottomrule
\end{tabular}
}
\subcaption{Energy consumption on ImageNet.\label{tab:energy-imagenet}}
\end{subtable}
\label{tab:section6}
\vskip -1cm
\end{table*}

As tabulated in~\cref{tab:energy-imagenet}, we calculate the energy consumption of BKDSNN according to the energy model in \cite{kundu2021hiresnn, zhou2022spikformer, hu2023advancing} and compare it with its ANN counterparts and SNNs trained by previous learning-based methods. 
In terms of the OPS, SNN variants are higher than ANNs, while ANNs use FLOPS as the operation counting metric. Nevertheless, since the energy per synaptic operation is more efficient than the operation in ANN.
For ResNet-50, BKDSNN consumes the lowest energy while optimally suppressing accuracy loss. For Sformer-8-768, our BKDSNN consumes more energy to increase firing rates for maintaining similar performance as SOTA methods for SNN training.

\section{Conclusion}
BKDSNN constructs a knowledge distillation based framework for learning-based SNN training, achieving promising performance on both static and neuromorphic datasets supported by comprehensive experiments. We expect that BKDSNN will contribute to future investigations on brain-inspired SNN, which can further narrow the performance gap between SNN and ANN. In this work, we mainly focus on learning-based SNN training methods due to their remarkable performance under ultra-low inference time-step. We anticipate to extend BKDSNN on conversion-based methods, which is expected to overcome the existing bottlenecks of conversion-based methods under ultra-low inference time-step. 

\section*{Acknowledgements}
This work is partially supported by National Key R\&D Program of China (2022YFB4500200), National Natural Science Foundation of China (No. 62102257),
Biren Technology–Shanghai Jiao Tong University Joint Laboratory Open Research Fund, Microsoft Research Asia Gift Fund, Lingang Laboratory Open Research Fund (No.LGQS-202202-11).

\bibliographystyle{splncs04}
\bibliography{main}

\begin{thebibliography}{10}
\providecommand{\url}[1]{\texttt{#1}}
\providecommand{\urlprefix}{URL }
\providecommand{\doi}[1]{https://doi.org/#1}

\bibitem{amir2017low}
Amir, A., Taba, B., Berg, D., Melano, T., McKinstry, J., Di~Nolfo, C., Nayak, T., Andreopoulos, A., Garreau, G., Mendoza, M., et~al.: A low power, fully event-based gesture recognition system. In: Proceedings of the IEEE conference on computer vision and pattern recognition. pp. 7243--7252 (2017)

\bibitem{bu2023optimal}
Bu, T., Fang, W., Ding, J., Dai, P., Yu, Z., Huang, T.: Optimal ann-snn conversion for high-accuracy and ultra-low-latency spiking neural networks. arXiv preprint arXiv:2303.04347  (2023)

\bibitem{cao2015spiking}
Cao, Y., Chen, Y., Khosla, D.: Spiking deep convolutional neural networks for energy-efficient object recognition. International Journal of Computer Vision  \textbf{113},  54--66 (2015)

\bibitem{cermelli2020modeling}
Cermelli, F., Mancini, M., Bulo, S.R., Ricci, E., Caputo, B.: Modeling the background for incremental learning in semantic segmentation. In: Proceedings of the IEEE/CVF Conference on Computer Vision and Pattern Recognition. pp. 9233--9242 (2020)

\bibitem{chen2019new}
Chen, L., Yu, C., Chen, L.: A new knowledge distillation for incremental object detection. In: 2019 International Joint Conference on Neural Networks (IJCNN). pp.~1--7. IEEE (2019)

\bibitem{davies2018loihi}
Davies, M., Srinivasa, N., Lin, T.H., Chinya, G., Cao, Y., Choday, S.H., Dimou, G., Joshi, P., Imam, N., Jain, S., et~al.: Loihi: A neuromorphic manycore processor with on-chip learning. Ieee Micro  \textbf{38}(1),  82--99 (2018)

\bibitem{deng2009imagenet}
Deng, J., Dong, W., Socher, R., Li, L.J., Li, K., Fei-Fei, L.: Imagenet: A large-scale hierarchical image database. In: 2009 IEEE conference on computer vision and pattern recognition. pp. 248--255. Ieee (2009)

\bibitem{deng2021optimal}
Deng, S., Gu, S.: Optimal conversion of conventional artificial neural networks to spiking neural networks. arXiv preprint arXiv:2103.00476  (2021)

\bibitem{deng2022temporal}
Deng, S., Li, Y., Zhang, S., Gu, S.: Temporal efficient training of spiking neural network via gradient re-weighting. arXiv preprint arXiv:2202.11946  (2022)

\bibitem{diehl2015fast}
Diehl, P., Neil, D., Binas, J., Cook, M., Liu, S.C., Pfeiffer, M.: Fast-classifying, high-accuracy spiking deep networks through weight and threshold balancing (07 2015). \doi{10.1109/IJCNN.2015.7280696}

\bibitem{ding2022snn}
Ding, J., Bu, T., Yu, Z., Huang, T., Liu, J.: Snn-rat: Robustness-enhanced spiking neural network through regularized adversarial training. Advances in Neural Information Processing Systems  \textbf{35},  24780--24793 (2022)

\bibitem{dosovitskiy2020image}
Dosovitskiy, A., Beyer, L., Kolesnikov, A., Weissenborn, D., Zhai, X., Unterthiner, T., Dehghani, M., Minderer, M., Heigold, G., Gelly, S., et~al.: An image is worth 16x16 words: Transformers for image recognition at scale. arXiv preprint arXiv:2010.11929  (2020)

\bibitem{spikingjelly}
Fang, W., Chen, Y., Ding, J., Yu, Z., Masquelier, T., Chen, D., Huang, L., Zhou, H., Li, G., Tian, Y.: Spikingjelly: An open-source machine learning infrastructure platform for spike-based intelligence. Science Advances  \textbf{9}(40),  eadi1480 (2023). \doi{10.1126/sciadv.adi1480}, \url{https://www.science.org/doi/abs/10.1126/sciadv.adi1480}

\bibitem{fang2021deep}
Fang, W., Yu, Z., Chen, Y., Huang, T., Masquelier, T., Tian, Y.: Deep residual learning in spiking neural networks. Advances in Neural Information Processing Systems  \textbf{34},  21056--21069 (2021)

\bibitem{garg2021dct}
Garg, I., Chowdhury, S.S., Roy, K.: Dct-snn: Using dct to distribute spatial information over time for low-latency spiking neural networks. In: Proceedings of the IEEE/CVF International Conference on Computer Vision. pp. 4671--4680 (2021)

\bibitem{gerstner2002spiking}
Gerstner, W., Kistler, W.M.: Spiking neuron models: Single neurons, populations, plasticity. Cambridge university press (2002)

\bibitem{han2020deep}
Han, B., Roy, K.: Deep spiking neural network: Energy efficiency through time based coding. In: European Conference on Computer Vision. pp. 388--404. Springer (2020)

\bibitem{han2020rmp}
Han, B., Srinivasan, G., Roy, K.: Rmp-snn: Residual membrane potential neuron for enabling deeper high-accuracy and low-latency spiking neural network. In: Proceedings of the IEEE/CVF conference on computer vision and pattern recognition. pp. 13558--13567 (2020)

\bibitem{he2016deep}
He, K., Zhang, X., Ren, S., Sun, J.: Deep residual learning for image recognition. In: Proceedings of the IEEE conference on computer vision and pattern recognition. pp. 770--778 (2016)

\bibitem{hinton2015distilling}
Hinton, G., Vinyals, O., Dean, J.: Distilling the knowledge in a neural network. arXiv preprint arXiv:1503.02531  (2015)

\bibitem{hong2023lasnn}
Hong, D., Shen, J., Qi, Y., Wang, Y.: Lasnn: Layer-wise ann-to-snn distillation for effective and efficient training in deep spiking neural networks. arXiv preprint arXiv:2304.09101  (2023)

\bibitem{2017CIFAR10}
Hongmin, L., Hanchao, L., Xiangyang, J., Guoqi, L., Luping, S.: Cifar10-dvs: An event-stream dataset for object classification. Frontiers in Neuroscience  \textbf{11} (2017)

\bibitem{hu2020spiking}
Hu, Y., Tang, H., Pan, G.: Spiking deep residual network (2020)

\bibitem{hu2023fast}
Hu, Y., Zheng, Q., Jiang, X., Pan, G.: Fast-snn: Fast spiking neural network by converting quantized ann. arXiv preprint arXiv:2305.19868  (2023)

\bibitem{hu2023advancing}
Hu, Y., Deng, L., Wu, Y., Yao, M., Li, G.: Advancing spiking neural networks towards deep residual learning (2023)

\bibitem{khan2020survey}
Khan, A., Sohail, A., Zahoora, U., Qureshi, A.S.: A survey of the recent architectures of deep convolutional neural networks. Artificial intelligence review  \textbf{53},  5455--5516 (2020)

\bibitem{kheradpisheh2018stdp}
Kheradpisheh, S.R., Ganjtabesh, M., Thorpe, S.J., Masquelier, T.: Stdp-based spiking deep convolutional neural networks for object recognition. Neural Networks  \textbf{99},  56--67 (2018)

\bibitem{krizhevsky2009learning}
Krizhevsky, A., Hinton, G., et~al.: Learning multiple layers of features from tiny images  (2009)

\bibitem{kundu2021hiresnn}
Kundu, S., Pedram, M., Beerel, P.A.: Hire-snn: Harnessing the inherent robustness of energy-efficient deep spiking neural networks by training with crafted input noise (2021)

\bibitem{kushawaha2021distilling}
Kushawaha, R.K., Kumar, S., Banerjee, B., Velmurugan, R.: Distilling spikes: Knowledge distillation in spiking neural networks. In: 2020 25th International Conference on Pattern Recognition (ICPR). pp. 4536--4543. IEEE (2021)

\bibitem{lecun2015deep}
LeCun, Y., Bengio, Y., Hinton, G.: Deep learning. nature  \textbf{521}(7553),  436--444 (2015)

\bibitem{lee2021energy}
Lee, D., Park, S., Kim, J., Doh, W., Yoon, S.: Energy-efficient knowledge distillation for spiking neural networks. arXiv preprint arXiv:2106.07172  (2021)

\bibitem{li2022quantization}
Li, C., Ma, L., Furber, S.: Quantization framework for fast spiking neural networks. Frontiers in Neuroscience  \textbf{16},  918793 (2022)

\bibitem{li2021free}
Li, Y., Deng, S., Dong, X., Gong, R., Gu, S.: A free lunch from ann: Towards efficient, accurate spiking neural networks calibration. In: International conference on machine learning. pp. 6316--6325. PMLR (2021)

\bibitem{liu2021sstdp}
Liu, F., Zhao, W., Chen, Y., Wang, Z., Yang, T., Li, J.: Sstdp: Supervised spike timing dependent plasticity for efficient spiking neural network training. Frontiers in neuroscience p.~1413 (2021)

\bibitem{liu2019structured}
Liu, Y., Chen, K., Liu, C., Qin, Z., Luo, Z., Wang, J.: Structured knowledge distillation for semantic segmentation. In: Proceedings of the IEEE/CVF conference on computer vision and pattern recognition. pp. 2604--2613 (2019)

\bibitem{maass1997networks}
Maass, W.: Networks of spiking neurons: the third generation of neural network models. Neural networks  \textbf{10}(9),  1659--1671 (1997)

\bibitem{merolla2014million}
Merolla, P.A., Arthur, J.V., Alvarez-Icaza, R., Cassidy, A.S., Sawada, J., Akopyan, F., Jackson, B.L., Imam, N., Guo, C., Nakamura, Y., et~al.: A million spiking-neuron integrated circuit with a scalable communication network and interface. Science  \textbf{345}(6197),  668--673 (2014)

\bibitem{neftci2019surrogate}
Neftci, E.O., Mostafa, H., Zenke, F.: Surrogate gradient learning in spiking neural networks: Bringing the power of gradient-based optimization to spiking neural networks. IEEE Signal Processing Magazine  \textbf{36}(6),  51--63 (2019)

\bibitem{orchard2015converting}
Orchard, G., Jayawant, A., Cohen, G.K., Thakor, N.: Converting static image datasets to spiking neuromorphic datasets using saccades. Frontiers in neuroscience  \textbf{9}, ~437 (2015)

\bibitem{ostojic2014two}
Ostojic, S.: Two types of asynchronous activity in networks of excitatory and inhibitory spiking neurons. Nature neuroscience  \textbf{17}(4),  594--600 (2014)

\bibitem{peng2021sid}
Peng, C., Zhao, K., Maksoud, S., Li, M., Lovell, B.C.: Sid: Incremental learning for anchor-free object detection via selective and inter-related distillation. Computer vision and image understanding  \textbf{210},  103229 (2021)

\bibitem{ranftl2021vision}
Ranftl, R., Bochkovskiy, A., Koltun, V.: Vision transformers for dense prediction. In: Proceedings of the IEEE/CVF international conference on computer vision. pp. 12179--12188 (2021)

\bibitem{roy2019towards}
Roy, K., Jaiswal, A., Panda, P.: Towards spike-based machine intelligence with neuromorphic computing. Nature  \textbf{575}(7784),  607--617 (2019)

\bibitem{rueckauer2017conversion}
Rueckauer, B., Lungu, I.A., Hu, Y., Pfeiffer, M., Liu, S.C.: Conversion of continuous-valued deep networks to efficient event-driven networks for image classification. Frontiers in neuroscience  \textbf{11}, ~682 (2017)

\bibitem{Selvaraju_2019}
Selvaraju, R.R., Cogswell, M., Das, A., Vedantam, R., Parikh, D., Batra, D.: Grad-cam: Visual explanations from deep networks via gradient-based localization. International Journal of Computer Vision  \textbf{128}(2),  336–359 (Oct 2019). \doi{10.1007/s11263-019-01228-7}, \url{http://dx.doi.org/10.1007/s11263-019-01228-7}

\bibitem{sengupta2019going}
Sengupta, A., Ye, Y., Wang, R., Liu, C., Roy, K.: Going deeper in spiking neural networks: Vgg and residual architectures. Frontiers in neuroscience  \textbf{13}, ~95 (2019)

\bibitem{sironi2018hats}
Sironi, A., Brambilla, M., Bourdis, N., Lagorce, X., Benosman, R.: Hats: Histograms of averaged time surfaces for robust event-based object classification. In: Proceedings of the IEEE conference on computer vision and pattern recognition. pp. 1731--1740 (2018)

\bibitem{sun2019patient}
Sun, S., Cheng, Y., Gan, Z., Liu, J.: Patient knowledge distillation for bert model compression. arXiv preprint arXiv:1908.09355  (2019)

\bibitem{takuya2021training}
Takuya, S., Zhang, R., Nakashima, Y.: Training low-latency spiking neural network through knowledge distillation. In: 2021 IEEE Symposium in Low-Power and High-Speed Chips (COOL CHIPS). pp.~1--3. IEEE (2021)

\bibitem{tan2022multi}
Tan, G., Wang, Y., Han, H., Cao, Y., Wu, F., Zha, Z.J.: Multi-grained spatio-temporal features perceived network for event-based lip-reading. In: Proceedings of the IEEE/CVF Conference on Computer Vision and Pattern Recognition. pp. 20094--20103 (2022)

\bibitem{wu2019direct}
Wu, Y., Deng, L., Li, G., Zhu, J., Xie, Y., Shi, L.: Direct training for spiking neural networks: Faster, larger, better. In: Proceedings of the AAAI conference on artificial intelligence. vol.~33, pp. 1311--1318 (2019)

\bibitem{xu2023constructing}
Xu, Q., Li, Y., Shen, J., Liu, J.K., Tang, H., Pan, G.: Constructing deep spiking neural networks from artificial neural networks with knowledge distillation. In: Proceedings of the IEEE/CVF Conference on Computer Vision and Pattern Recognition. pp. 7886--7895 (2023)

\bibitem{xu2022hierarchical}
Xu, Q., Li, Y., Shen, J., Zhang, P., Liu, J.K., Tang, H., Pan, G.: Hierarchical spiking-based model for efficient image classification with enhanced feature extraction and encoding. IEEE Transactions on Neural Networks and Learning Systems  (2022)

\bibitem{xu2022delving}
Xu, Z., Zhang, M., Hou, J., Gong, X., Wen, C., Wang, C., Zhang, J.: Delving into transformer for incremental semantic segmentation (2022)

\bibitem{yang2022focal}
Yang, Z., Li, Z., Jiang, X., Gong, Y., Yuan, Z., Zhao, D., Yuan, C.: Focal and global knowledge distillation for detectors. In: Proceedings of the IEEE/CVF Conference on Computer Vision and Pattern Recognition. pp. 4643--4652 (2022)

\bibitem{yao2023spike}
Yao, M., Hu, J., Hu, T., Xu, Y., Zhou, Z., Tian, Y., Bo, X., Li, G.: Spike-driven transformer v2: Meta spiking neural network architecture inspiring the design of next-generation neuromorphic chips. In: The Twelfth International Conference on Learning Representations (2023)

\bibitem{zenke2015diverse}
Zenke, F., Agnes, E.J., Gerstner, W.: Diverse synaptic plasticity mechanisms orchestrated to form and retrieve memories in spiking neural networks. Nature communications  \textbf{6}(1), ~6922 (2015)

\bibitem{zenke2021remarkable}
Zenke, F., Vogels, T.P.: The remarkable robustness of surrogate gradient learning for instilling complex function in spiking neural networks. Neural computation  \textbf{33}(4),  899--925 (2021)

\bibitem{zheng2021going}
Zheng, H., Wu, Y., Deng, L., Hu, Y., Li, G.: Going deeper with directly-trained larger spiking neural networks. In: Proceedings of the AAAI conference on artificial intelligence. vol.~35, pp. 11062--11070 (2021)

\bibitem{zhou2023spikingformer}
Zhou, C., Yu, L., Zhou, Z., Zhang, H., Ma, Z., Zhou, H., Tian, Y.: Spikingformer: Spike-driven residual learning for transformer-based spiking neural network. arXiv preprint arXiv:2304.11954  (2023)

\bibitem{zhou2023enhancing}
Zhou, C., Zhang, H., Zhou, Z., Yu, L., Ma, Z., Zhou, H., Fan, X., Tian, Y.: Enhancing the performance of transformer-based spiking neural networks by improved downsampling with precise gradient backpropagation. arXiv preprint arXiv:2305.05954  (2023)

\bibitem{zhou2021deepvit}
Zhou, D., Kang, B., Jin, X., Yang, L., Lian, X., Jiang, Z., Hou, Q., Feng, J.: Deepvit: Towards deeper vision transformer. arXiv preprint arXiv:2103.11886  (2021)

\bibitem{zhou2021rethinking}
Zhou, H., Song, L., Chen, J., Zhou, Y., Wang, G., Yuan, J., Zhang, Q.: Rethinking soft labels for knowledge distillation: A bias-variance tradeoff perspective. arXiv preprint arXiv:2102.00650  (2021)

\bibitem{zhou2022spikformer}
Zhou, Z., Zhu, Y., He, C., Wang, Y., Yan, S., Tian, Y., Yuan, L.: Spikformer: When spiking neural network meets transformer (2022)

\end{thebibliography}

\clearpage
\setcounter{page}{1}
\setcounter{section}{0}
\setcounter{table}{0}
\setcounter{figure}{0}
\setcounter{equation}{0}
\renewcommand{\thesection}{A\arabic{section}}
\renewcommand{\thetable}{A\arabic{table}}
\renewcommand{\thefigure}{A\arabic{figure}}
\renewcommand{\theequation}{A\arabic{equation}}
\begin{center}
    \begin{Large}
    \textbf{Supplementary}
    \end{Large}
\end{center}

\section{Definition of Time-Step.} 

In spiking neural network (SNN), \textit{time-step} refers to a discrete unit of time used in the simulation of the network dynamics~\cite{bu2023optimal, diehl2015fast,li2022quantization, rueckauer2017conversion}. Unlike the traditional artificial neural network (ANN), which often operates in a continuous manner, SNN models the behavior of biological neurons by considering the timing (time-stamp) of spikes. The simulation of SNN typically involves breaking down time into discrete steps, and for each time-step, the network processes incoming signals, updates neuron membrane potentials, and generates spikes based on certain rules and conditions. The time-step is a fundamental parameter that influences the temporal resolution of the simulation and is crucial for capturing the dynamics of spiking neurons. The choice of an appropriate time-step is essential for \emph{balancing computational efficiency and accuracy in simulating the behavior of the network}. With greater time-step, SNN can provides a more accurate representation of the temporal dynamics but the inference consumes more energy. With smaller time-step, on the other hand, it can speed up simulations but may lead to a loss of precision (\ie, inference accuracy).

\begin{algorithm}[t]
  \caption{Training Algorithm of BKDSNN}
  \label{alg:bkdsnn}
\textbf{Input}: Dataset \textit{D}.\\ 
\textbf{Model}: Teacher ANN model $\mathcal{M}_\text{ANN}(x;\Tilde{\tW})$ with pretrained weight $\Tilde{\tW}$; Student SNN model $\mathcal{M}_\text{SNN}(x;\tW)$ with initial weight $\tW$; Blurred restoration block $\mathcal{G}$. \\ 
\textbf{Parameter}: Temperature $\tau$ and weight $w_\text{LD}$ for $\mathcal{L}_\text{LD}$; Blurred ratio $\lambda$ and weight $w_\text{BKD}$ for $\mathcal{L}_\text{BKD}$.\\
\textbf{Output}: Student SNN model $\mathcal{M}_\text{SNN}(x;\tW)$.\\
\begin{algorithmic}[1]
\FOR{$i=1$ to $e$}
\FOR{$x, \hat{y}$ in \textit{D}}
\STATE  \texttt{\# forward propagation}
\STATE $\vy^\text{tea}, \tF^\text{tea} = \mathcal{M}_\text{ANN}(x)$;
\STATE $\vy^\text{stu}, \tF^\text{stu} = \mathcal{M}_\text{SNN}(x)$;
\STATE \#\textit{calculate} $\mathcal{L}_\text{LD}$
\STATE calculate $\vq^\text{stu}$ and $\vq^\text{tea}$ using~\cref{eq:flatten};
\STATE $\mathcal{L}_\text{LD}=\tau^2 \cdot \mathcal{L}_\text { CrossEntropy }\left(\vq^\text{stu}, \vq^\text{tea}\right)$;
\STATE \texttt{\# calculate} $\mathcal{L}_\text{BKD}$
\STATE calculate $\hat{\tF}^\text{stu}$ using~\cref{eq:bm}, \cref{eq:mean} and~\cref{eq:restoration};
\STATE $\mathcal{L}_\text{BKD}= \displaystyle\sum_{i}^{b}\sum_{j}^{c^\text{stu}}\sum_{k}^{h}\sum_{l}^{w}\left(\hat{\tF}_{i, j, k, l}^\text{stu}-\tF_{i, j, k, l}^\text{tea}\right)^2$;
\STATE \texttt{\# calculate total loss} $\mathcal{L}$
\STATE $\mathcal{L}_\text{MD}= w_\text{LD} \cdot \mathcal{L}_\text{LD} + w_\text{BKD} \cdot \mathcal{L}_\text{BKD}$;
\STATE $\mathcal{L} = \mathcal{L}_\text { CrossEntropy }\left(\vy^\text{stu}, \hat{\vy}\right) + \mathcal{L}_\text{MD} = \mathcal{L}_\text { task} + \mathcal{L}_\text{MD}$;
\STATE \texttt{\# backward propagation}
\STATE calculate the gradients $\frac{\partial\mathcal{L}}{\partial\tW}$;
\STATE update $\tW$;
\ENDFOR
\ENDFOR
\end{algorithmic}
\label{alg:alg}
\end{algorithm}

\section{Algorithm for BKDSNN.} The process of our algorithm is specified in~\cref{alg:alg}. In each training iteration, we first calculate the logits-based distillation loss $\mathcal{L}_\text{LD}$, using output logits $\vy^\text{tea}$ and $\vy^\text{stu}$.
Then, we extract the intermediate feature $\tF^\text{tea}$ and $\tF^\text{stu}$ in front of the last layer from the teacher ANN $\mathcal{M}_\text{ANN}$ and the student SNN $\mathcal{M}_\text{SNN}$ respectively. Afterwards, we leverage $\tF^\text{tea}$, $\tF^\text{stu}$ together with the blurred restoration block $\mathcal{G}$ and the blurred ratio $\lambda$ to calculate a blurred knowledge distillation loss $\mathcal{L}_\text{BKD}$. 
Combined with the cross entropy task loss between $\vy^\text{tea}$ and ground truth label $\hat{\vy}$, we obtain the total loss $\mathcal{L}$ which can be used with back-propagation through time (BPTT)~\cite{wu2019direct, zenke2021remarkable} algorithm to update $\tW$ in $\mathcal{M}_\text{SNN}$.

\section{Ablation Study on CIFAR.} Ablation results on CIFAR are shown in~\cref{tab:ablation-cifar-t,tab:ablation-cifar-t-400,tab:ablation-cifar-c}, where the introduction of logits-based distillation loss $\mathcal{L}_\text{LD}$, blurred knowledge distillation $\mathcal{L}_\text{BKD}$ and mixed distillation $\mathcal{L}_\text{MD}$ significantly improve the performance, compared with methods without knowledge distillation~(denoted as w/o \textit{KD}). However, as shown in~\cref{tab:ablation-cifar-t}, the introduction of $\mathcal{L}_\text{LD}$ and $\mathcal{L}_\text{BKD}$ are more effective than $\mathcal{L}_\text{MD}$ on CIFAR10 with Sformer-2-384 and Sformer-4-384. \emph{Limited training epochs on a simple dataset lead to the experimental results above}. Therefore, we increase the training epochs from 300 to 400 for more distinct results. As shown in~\cref{tab:ablation-cifar-t-400}, the introduction of $\mathcal{L}_\text{MD}$ achieves maximized accuracy boost under all settings on CIFAR, leading to state-of-the-art performance on CIFAR~(96.37\% on CIFAR10 and 81.63\% on CIFAR100). Along with the experimental results on CIFAR with CNN-based models in~\cref{tab:ablation-cifar-c}, the effectiveness of our BKDSNN is further verified.

\begin{figure}[t]
    \centering
    \includegraphics[width=0.4\linewidth]{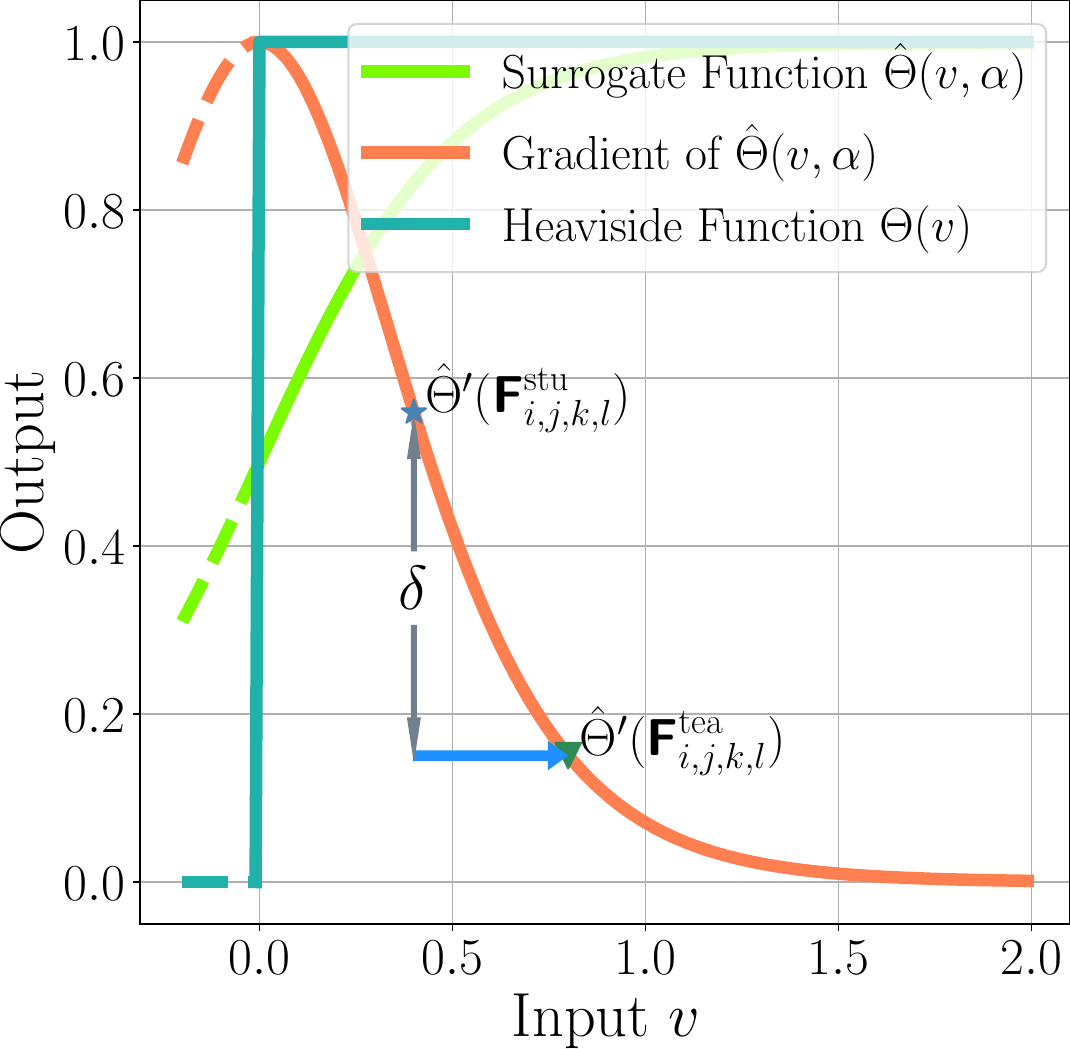}
\caption{Illustration of gradient correction with BKD for IF neuron.}
	\label{fig:vissurrogate}
\end{figure}

\begin{table}[t]
\caption{Ablation on CIFAR with Transformer-based models after 300 epochs.}
\centering
\resizebox{0.7\textwidth}{!}{
\begin{tabular}{lccccccc}
\toprule
\multirow{3}{*}{Method}   &       & \multicolumn{3}{c}{CIFAR10}     & \multicolumn{3}{c}{CIFAR100}    \\
 & \multirow{2}{*}{\begin{tabular}[c]{@{}c@{}}Student\\ SNN\end{tabular}} & \multicolumn{2}{c}{Teacher ANN} & \multirow{2}{*}{\begin{tabular}[c]{@{}c@{}}SNN\\ Acc\end{tabular}} & \multicolumn{2}{c}{Teacher ANN} & \multirow{2}{*}{\begin{tabular}[c]{@{}c@{}}SNN\\ Acc\end{tabular}} \\
 &       & model  & Acc &   & model  & Acc &   \\ \midrule
\multirow{3}{*}{\begin{tabular}[c]{@{}l@{}}w/o\\ \textit{KD}\end{tabular}} & \textit{S}-4-256~\cite{xu2023constructing} & -    & - & 94.94 & -    & - & 78.19 \\
 & \textit{S}-2-384~\cite{xu2023constructing} & -    & - & 95.54 & -    & - & 78.87 \\
 & \textit{S}-4-384~\cite{xu2023constructing} & -    & - & 95.81 & -    & - & 79.98 \\ \midrule
\multirow{3}{*}{$+\mathcal{L}_\text{LD}$}    & \textit{S}-4-256 & \multirow{3}{*}{ViT-S/16} & \multirow{3}{*}{96.75} & 95.08(+0.14)   & \multirow{3}{*}{ViT-S/16} & \multirow{3}{*}{82.22} & 78.73(+0.54)   \\
 & \textit{S}-2-384 &      &   & \textbf{95.96(+0.42)}   &      &   & 80.02(+1.15)   \\
 & \textit{S}-4-384 &      &   & \baseline{96.07(+0.26)}   &      &   & 80.82(+0.84)   \\ \midrule
\multirow{3}{*}{$+\mathcal{L}_\text{BKD}$}    & \textit{S}-4-256 & \multirow{3}{*}{ViT-S/16} & \multirow{3}{*}{96.75} & \baseline{95.22(+0.28)}   & \multirow{3}{*}{ViT-S/16} & \multirow{3}{*}{82.22} & \baseline{79.22(+1.03)}   \\
 & \textit{S}-2-384 &      &   & 95.75(+0.21)   &      &   & \baseline{80.26(+1.39)}   \\
 & \textit{S}-4-384 &      &   & \textbf{96.12(+0.31)}  &      &   & \baseline{81.18(+1.20) }  \\ \midrule
\multirow{3}{*}{$+\mathcal{L}_\text{MD}$}    & \textit{S}-4-256 & \multirow{3}{*}{ViT-S/16} & \multirow{3}{*}{96.75} & \textbf{95.29(+0.35)}  & \multirow{3}{*}{ViT-S/16} & \multirow{3}{*}{82.22} & \textbf{79.41(+1.22)}  \\
 & \textit{S}-2-384 &      &   & \baseline{95.90(+0.36)}  &      &   & \textbf{80.63(+1.76)}  \\
 & \textit{S}-4-384 &      &   & 96.06(+0.25)   &      &   & \textbf{81.26(+1.28)}  \\ \bottomrule
\end{tabular}
}
\label{tab:ablation-cifar-t}
\end{table}

\begin{table}[t]
\caption{Ablation on CIFAR with Transformer-based models after 400 epochs.}
\centering
\resizebox{0.7\textwidth}{!}{
\begin{tabular}{lccccccc}
\toprule
\multirow{3}{*}{Method}      &   & \multicolumn{3}{c}{CIFAR10}      & \multicolumn{3}{c}{CIFAR100}     \\
     & \multirow{2}{*}{\begin{tabular}[c]{@{}c@{}}Student\\ SNN\end{tabular}} & \multicolumn{2}{c}{Teacher ANN} & \multirow{2}{*}{\begin{tabular}[c]{@{}c@{}}SNN\\ Acc\end{tabular}} & \multicolumn{2}{c}{Teacher ANN} & \multirow{2}{*}{\begin{tabular}[c]{@{}c@{}}SNN\\ Acc\end{tabular}} \\
     &   & model  & Acc &       & model  & Acc &       \\ \midrule
\multirow{3}{*}{\begin{tabular}[c]{@{}l@{}}w/o\\ \textit{KD}\end{tabular}} & \textit{S}-4-256-400E~\cite{xu2023constructing} & -  & -   & 95.31 & -  & -   & 78.58 \\
     & \textit{S}-2-384-400E~\cite{xu2023constructing} & -  & -   & 95.75 & -  & -   & 79.17 \\
     & \textit{S}-4-384-400E~\cite{xu2023constructing} & -  & -   & 95.95 & -  & -   & 80.37 \\ \midrule
\multirow{3}{*}{$+\mathcal{L}_\text{LD}$} & \textit{S}-4-256-400E & \multirow{3}{*}{ViT-S/16} & \multirow{3}{*}{96.75} & \baseline{95.60(+0.29)}        & \multirow{3}{*}{ViT-S/16} & \multirow{3}{*}{82.22} & 79.54(+0.96)        \\
     & \textit{S}-2-384-400E & & & \baseline{95.95(+0.20)}        & & & 80.56(+1.39)        \\
     & \textit{S}-4-384-400E & & & \baseline{96.17(+0.22)}        & & & 81.25(+0.88)        \\ \midrule
\multirow{3}{*}{$+\mathcal{L}_\text{BKD}$} & \textit{S}-4-256-400E & \multirow{3}{*}{ViT-S/16} & \multirow{3}{*}{96.75} & 95.43(+0.12)        & \multirow{3}{*}{ViT-S/16} & \multirow{3}{*}{82.22} & \baseline{79.63(+1.05)}        \\
     & \textit{S}-2-384-400E & & & 95.93(+0.18)        & & & \baseline{80.83(+1.66)}        \\
     & \textit{S}-4-384-400E & & & 96.13(+0.18)        & & & \baseline{81.38(+1.01)}        \\ \midrule
\multirow{3}{*}{$+\mathcal{L}_\text{MD}$} & \textit{S}-4-256-400E & \multirow{3}{*}{ViT-S/16} & \multirow{3}{*}{96.75} & \textbf{95.76(+0.45)} & \multirow{3}{*}{ViT-S/16} & \multirow{3}{*}{82.22} & \textbf{79.86(+1.28)} \\
     & \textit{S}-2-384-400E & & & \textbf{96.18(+0.43)} & & & \textbf{80.83(+1.66)} \\
     & \textit{S}-4-384-400E & & & \textbf{96.37(+0.42)} & & & \textbf{81.63(+1.26)} \\ \bottomrule
\end{tabular}
}
\label{tab:ablation-cifar-t-400}
\end{table}

\begin{table}[t]
\caption{Ablation on CIFAR with CNN-based models.}
\centering
\resizebox{0.7\textwidth}{!}{
\begin{tabular}{lccccccc}
\toprule
\multirow{3}{*}{Method} & & \multicolumn{3}{c}{CIFAR10}   & \multicolumn{3}{c}{CIFAR100}  \\
    & \multirow{2}{*}{\begin{tabular}[c]{@{}c@{}}Student\\ SNN\end{tabular}} & \multicolumn{2}{c}{Teacher ANN} & \multirow{2}{*}{\begin{tabular}[c]{@{}c@{}}SNN\\ Acc\end{tabular}} & \multicolumn{2}{c}{Teacher ANN} & \multirow{2}{*}{\begin{tabular}[c]{@{}c@{}}SNN\\ Acc\end{tabular}} \\
    & & model  & Acc &      & model  & Acc &      \\ \midrule
\multirow{2}{*}{\begin{tabular}[c]{@{}l@{}}w/o\\ \textit{KD}\end{tabular}} & TET \textit{R}-19~\cite{deng2022temporal}  & -   & -   & 94.44 & -   & -   & 74.47 \\
    & SEW \textit{R}-20~\cite{fang2021deep}  & -   & -   & 89.07 & -   & -   & 60.16 \\ \hline
\multirow{2}{*}{$+\mathcal{L}_\text{LD}$}      & TET \textit{R}-19  & \textit{R}-19   & 95.60  & 94.49(+0.05)  & \textit{R}-19   & 79.78  & 74.58(+0.11)  \\
    & SEW \textit{R}-20  & \textit{R}-20   & 91.77  & 89.13(+0.06)  & \textit{R}-20   & 68.40  & 60.36(+0.20)  \\ \midrule
\multirow{2}{*}{$+\mathcal{L}_\text{BKD}$}      & TET \textit{R}-19  & \textit{R}-19   & 95.60  & \baseline{94.58(+0.14)}  & \textit{R}-19   & 79.78  & \baseline{74.82(+0.35)}  \\
    & SEW \textit{R}-20  & \textit{R}-20   & 91.77  & \baseline{89.21(+0.14)}  & \textit{R}-20   & 68.40  & \baseline{60.57(+0.41)}  \\ \midrule
\multirow{2}{*}{$+\mathcal{L}_\text{MD}$}      & TET \textit{R}-19  & \textit{R}-19   & 95.60  & \textbf{94.64(+0.20)}     & \textit{R}-19   & 79.78  & \textbf{74.95(+0.48)}     \\
    & SEW \textit{R}-20  & \textit{R}-20   & 91.77  & \textbf{89.29(+0.22)}     & \textit{R}-20   & 68.40  & \textbf{60.92(+0.76)}     \\ \bottomrule
\end{tabular}
}
\label{tab:ablation-cifar-c}
\end{table}

\begin{figure*}[t]
\centering
\subcaptionbox{Q, K, V and attention firing rate curve of Sformer-4-384 under multiple settings on CIFAR10.\label{fig:fr-cifar10}}{\includegraphics[width=\textwidth]{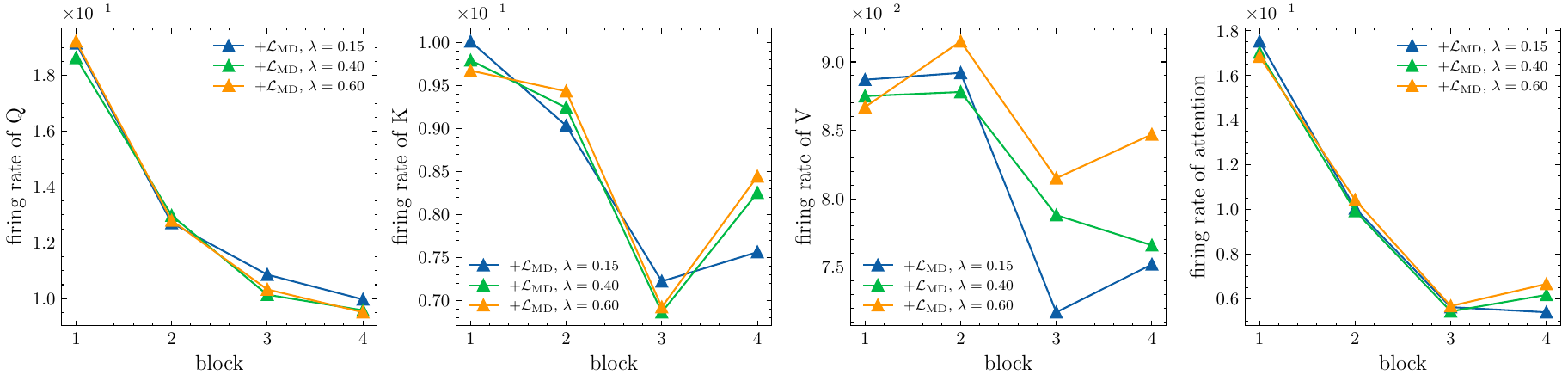}} \\
\subcaptionbox{Q, K, V and attention firing rate curve of Sformer-4-384 under multiple settings on CIFAR100.\label{fig:fr-cifar100}}{\includegraphics[width=\textwidth]{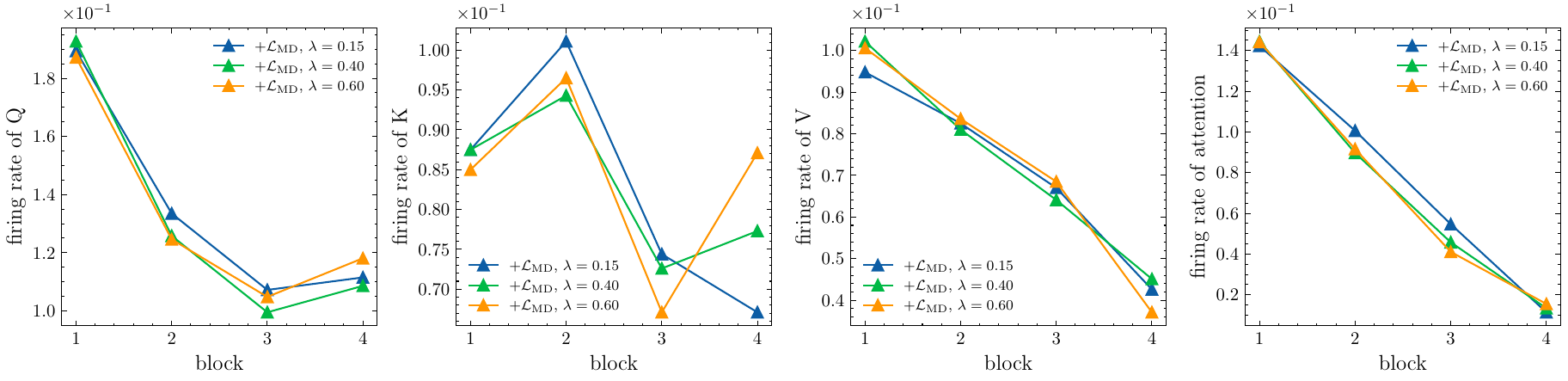}} \\
\subcaptionbox{Q, K, V and attention firing rate curve of Sformer with multiple feature dimension on ImageNet.}{\includegraphics[width=\textwidth]{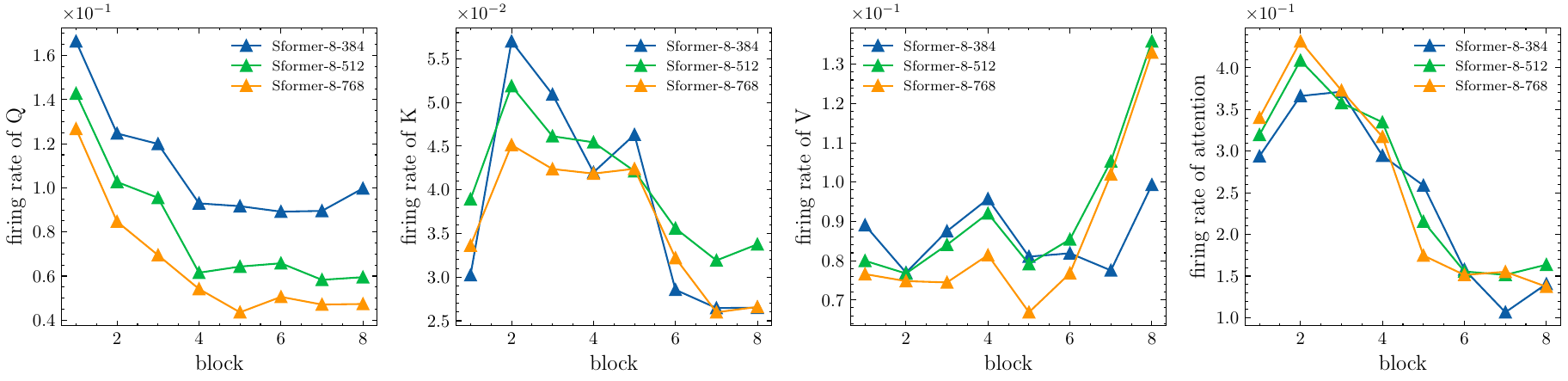}} \\

\caption{\textbf{Q, K, V and attention firing rate curve of Sformer on both CIFAR and ImageNet.} The top 2 rows are Q, K, V and attention firing rate curve with multiple blurred ratio $\lambda$ on CIFAR, it is obvious that higher blurred ratio $\lambda$ leads to higher firing rate. The bottom row is Q, K, V and attention firing rate curve of Sformer with multiple feature dimension on ImageNet, Sformer with higher dimension fires more spikes in attention map.}
\label{fig:qkva}
\end{figure*}

\section{Effectiveness of BKD on Gradient Correction. } We further analyse the advantages of proposed BKD on SNN training. In learning-based SNN training, spiking neuron is a vital component. As illustrated in \cref{eq:neuron}, spiking neuron leverages the heaviside function $\Theta\left(\cdot\right)$ to generate spikes. However, the gradient of heaviside function $\Theta\left(\cdot\right)$~(\cref{eq:neurongrad}) is not applicable to classical back-propagation based training as the existence of infinity value results in instability.
\begin{equation}
\dfrac{\partial \Theta (v)}{\partial v}
= \begin{cases}+\infty, & v = 0 \\ 0, & \text{otherwise}\end{cases}
\label{eq:neurongrad}
\end{equation}
Aiming at avoiding infinity value during back-propogation, \cite{neftci2019surrogate} proposed the surrogate gradient method which introduced a surrogate function to replace the Heaviside function. In this paper we use the sigmoid function $\hat{\Theta}(\cdot, \alpha)$~(\cref{eq:surrogate}) as the surrogate function due to its versatility and stability~\cite{zhou2022spikformer, zhou2023spikingformer, zhou2023enhancing}. 
\begin{equation}
\label{eq:surrogate}
\begin{gathered}
\hat{\Theta}\left(v, \alpha\right)= \frac{1}{1+e^{-\alpha v}} \\
\hat{\Theta}'\left(v, \alpha\right) = \alpha \hat{\Theta}\left(v, \alpha\right) - \alpha \hat{\Theta}^{2}\left(v, \alpha\right)
\end{gathered}
\end{equation}
where $\alpha$ refers to smoothing factor and we set it to 4.0 as previous works~\cite{zhou2022spikformer, zhou2023spikingformer, zhou2023enhancing}. Then we derive the gradient $\frac{\partial \mathcal{L}_\text{total}}{\partial \tF^\text{stu}_{i, j, k, l}}$ generated by $\mathcal{L}_\text{total}$ in \cref{eq:Lgrad} 
\begin{equation}
\begin{split}
\frac{\partial \mathcal{L}_\text{total}}{\partial \tF^\text{stu}_{i, j, k, l}} &= \frac{\partial \mathcal{L}_\text{CE}}{\partial \vy^\text{stu}}\frac{\partial \vy^\text{stu}}{\partial \tF^\text{stu}_{i, j, k, l}} + \frac{\partial \mathcal{L}_\text{BKD}}{\partial \hat{\tF}_{i, j, k, l}^\text{stu}}\frac{\partial \hat{\tF}_{i, j, k, l}^\text{stu}}{\partial \tF_{i, j, k, l}^\text{stu}} \\
&= \frac{\partial \mathcal{L}_\text{CE}}{\partial \vy^\text{stu}} \hat{\Theta}'\left(\tF^\text{stu}_{i, j, k, l}, \alpha\right) + \delta
\label{eq:Lgrad}
\end{split}
\end{equation}


As depicted in \cref{fig:vissurrogate}, $\delta$ in~\cref{eq:Lgrad} generated by $\mathcal{L}_\text{BKD}$ serves as a correction term which refines the surrogate gradient for more precise estimation in learning-based SNN training. The GradCAM visualization in~\cref{fig:visfmapcop} further collaborates that the introduction of $\mathcal{L}_\text{BKD}$ is beneficial for surrogate gradient estimation.

\section{Firing Rate of Q, K, V.} 
\begin{figure*}[h]
\centering
\subcaptionbox{Firing rate curve of Q in Sformer-8-768 with multiple methods on ImageNet.}{\includegraphics[width=\textwidth]{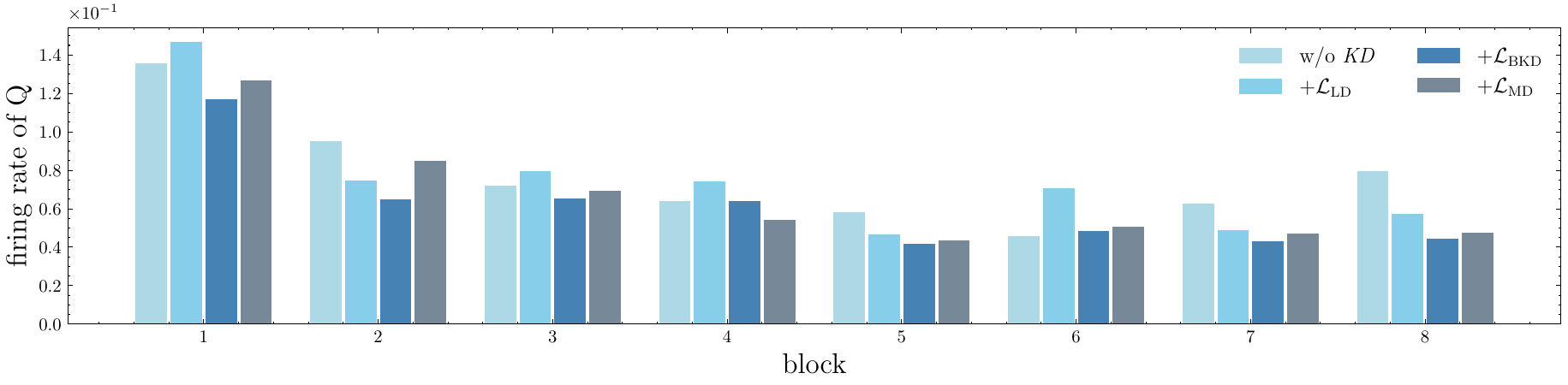}} \\
\subcaptionbox{Firing rate curve of K in Sformer-8-768 with multiple methods on ImageNet.}{\includegraphics[width=\textwidth]{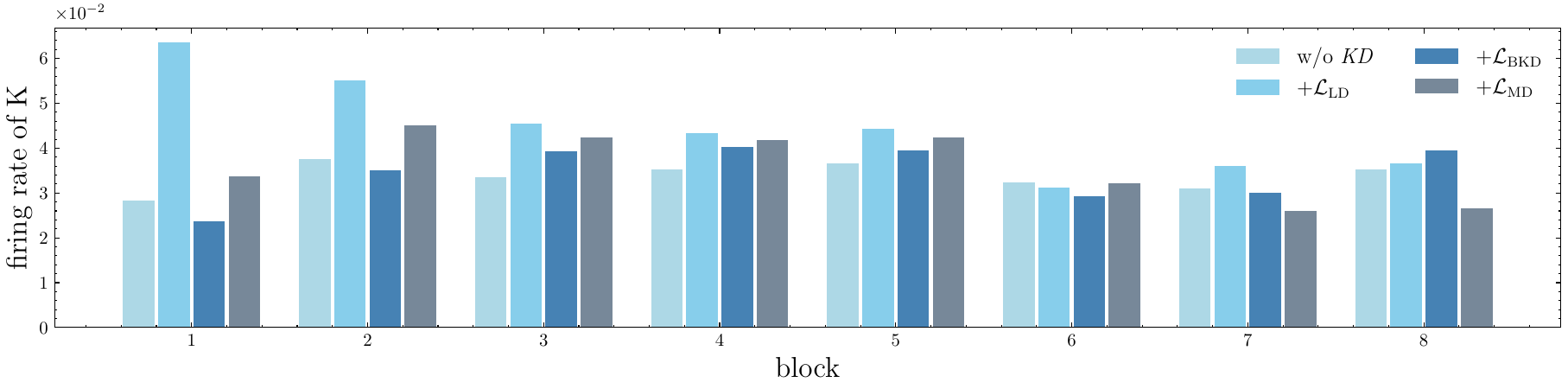}} \\
\subcaptionbox{Firing rate curve of V in Sformer-8-768 with multiple methods on ImageNet.}{\includegraphics[width=\textwidth]{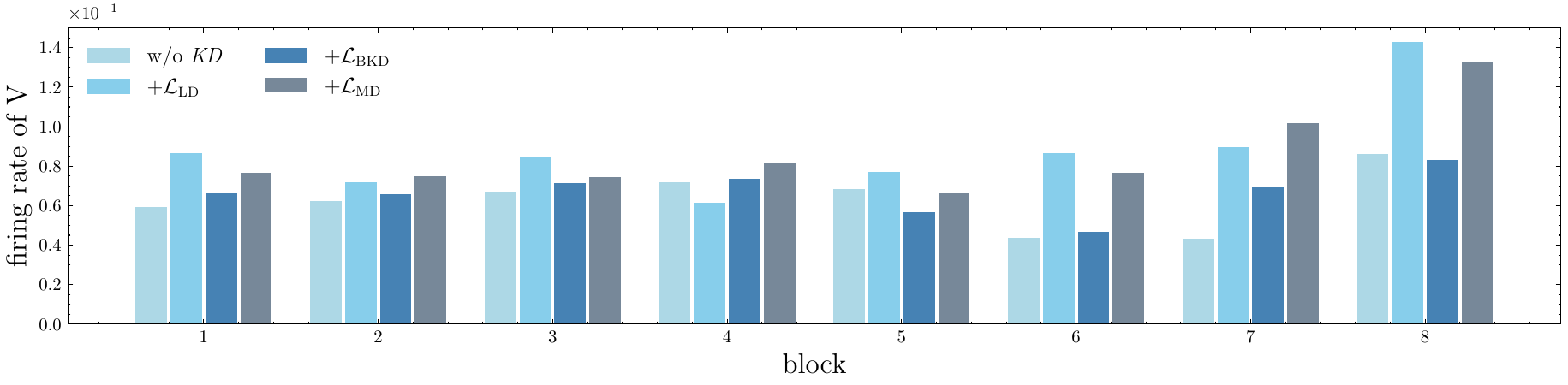}} \\
\subcaptionbox{Firing rate curve of attention in Sformer-8-768 with multiple methods on ImageNet.}{\includegraphics[width=\textwidth]{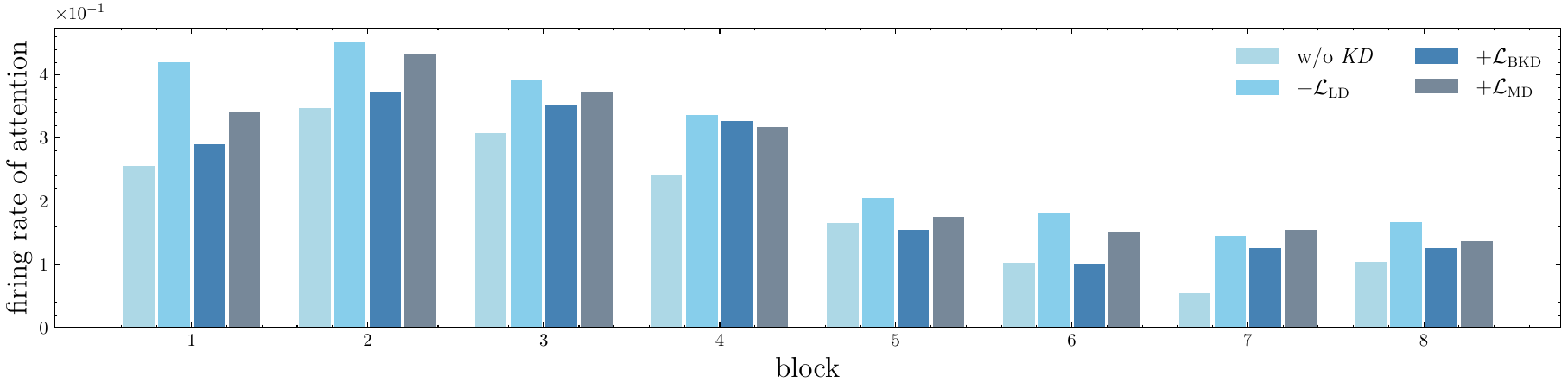}} \\

\caption{\textbf{Q, K, V and attention firing rate curve of Sformer-8-768 on ImageNet with different methods.} The introduction of distillation increases the firing rate of Q, K, V and attention compared with method without distillation~(denoted as w/o KD). Compared with logits-based distillation~(denoted as $+\mathcal{L}_\text{LD}$), our blurred knowledge distillation~(denoted as $+\mathcal{L}_\text{BKD}$) increases firing rate by a smaller margin while achieving higher accuracy improvement~(+0.60\%). Compared with method without distillation, mixed distillation~(denoted as $+\mathcal{L}_\text{MD}$) trades an acceptable firing rate increase for the maximized accuracy boost~(+2.29\%).}
\label{fig:qkva-768}
\end{figure*}

We plot Q, K, V and attention firing rate curve of Sformer-8-768 on ImageNet, with different methods in~\cref{fig:qkva-768}. The introduction of distillation increases the firing rate of Q, K, V and attention compared with method without distillation~(denoted as w/o KD). Compared with logits-based distillation~(denoted as $+\mathcal{L}_\text{LD}$), our blurred knowledge distillation~(denoted as $+\mathcal{L}_\text{BKD}$) increases the firing rate by a smaller margin while achieving higher accuracy improvement~(+0.60\%). Compared with method without distillation, mixed distillation~(denoted as $+\mathcal{L}_\text{MD}$) trades an acceptable firing rate increase for the maximized accuracy boost~(+2.29\%). The increment of firing rate leads to higher energy consumption which is verified in~\cref{tab:energy-imagenet}. However, with the significant boost on accuracy~(+2.29\%), such energy consumption increase~(+2.24\%) is acceptable.

\section{Energy Consumption Calculation.} 
We follow previous works~\cite{kundu2021hiresnn, zhou2022spikformer, hu2023advancing} to calculate the energy consumption of our BKDSNN and tabulate the results in~\cref{tab:energy-imagenet}. 
Calculating the theoretical energy consumption requires to count the synaptic operations $\operatorname{SOPS}$ as follows:
\begin{equation}
\operatorname{SOPS}(l)=r_\text{input} \times T \times \operatorname{FLOPS}(l)
\end{equation}
where $l$ is a block/layer in corresponding models, $r_\text{input}$ is the firing rate of the input spike train of the block/layer, $T$ is the simulation time-step of spike neuron. $\operatorname{FLOPS}(l)$ refers to the floating point operations of $l$ block/layer, which is the number of multiply-and-accumulate~(MAC) operations. $\operatorname{SOPS}(l)$ refers to the spike-based accumulate~(AC) operations of $l$ block/layer. Then, we follow the assumption in~\cite{zhou2022spikformer} that both MAC and AC operations are implemented on the 45nm hardware, where $E_\text{MAC}=4.6 \textrm{pJ}$ and $E_\text{AC}=0.9 \textrm{pJ}$. The theoretical energy consumption $E$ is defined as:
\begin{equation}
E=E_\text{MAC} \times \operatorname{FLOPS}(1) + E_\text{AC} \times \displaystyle\sum_{i=2}^{l} \operatorname{SOPS}(i)
\end{equation}
where $\operatorname{FLOPS}(1)$ refers to the first layer to encode static RGB images into the form of spikes.

\section{Experiments on Multiple Neuromorphic Datasets}
\vskip -1cm
\begin{table}[h]
\centering
\caption{Experiments on Multiple Neuromorphic Datasets.\label{tab:nd}}
\vskip -0.2cm
\resizebox{0.7\textwidth}{!}{
\begin{tabular}{lcccccccc}
\toprule
\multirow{2}{*}{Method}        & \multicolumn{2}{c}{DVS-Gesture}       & \multicolumn{2}{c}{NCaltech-101}        & \multicolumn{2}{c}{N-CARS}    & \multicolumn{2}{c}{DVSLip}  \\
  & Acc & T & Acc   & T & Acc   & T & Acc & T \\ \midrule
Sformer+CML & 98.6  & 16 & 69.91 & 16 & 99.31 & 16 & 54.93 &  15  \\ \midrule
Sformer+CML  & 96.3  & 4  & 69.12 & 4  & 98.42 & 4  & 49.84 & 10 \\
KDSNN       & 96.4  & 4  & 69.16 & 4  & \cellcolor{baselinecolor}98.51 & 4  & \cellcolor{baselinecolor}51.06 & 10  \\
LaSNN       & \cellcolor{baselinecolor}96.7  & 4  & \cellcolor{baselinecolor}69.21 & 4  & 98.44 & 4  &  50.82   &  10  \\
\begin{tabular}[c]{@{}l@{}}BKDSNN\\ (ours)\end{tabular}      & \textbf{\begin{tabular}[c]{@{}c@{}}97.8\\ (+1.1)\end{tabular}} & 4  & \textbf{\begin{tabular}[c]{@{}c@{}}69.68\\ (+0.47)\end{tabular}} & 4  & \textbf{\begin{tabular}[c]{@{}c@{}}99.02\\ (+0.51)\end{tabular}} & 4  &  \textbf{\begin{tabular}[c]{@{}c@{}}52.96\\ (+1.90)\end{tabular}}   & 10   \\ \bottomrule
\end{tabular}
}
\vskip -0.5cm
\end{table}

As tabulated in~\cref{tab:nd}, we extend our BKDSNN on datasets of DVS-Gesture~\cite{amir2017low}, NCaltech-101~\cite{orchard2015converting}, N-CARS~\cite{sironi2018hats} and DVSLip~\cite{tan2022multi}, BKDSNN achieves state-of-the-arts results on all datasets above, which further verifies the practical utility of BKDSNN on neuromorphic datasets.

\section{Experiments on Object Detection and Semantic Segmentation}
\vskip -1cm
\begin{table}[h]
\centering
\caption{\footnotesize Experiments with M-Sformer~(Meta-SpikeFormer) on object detection and semantic segmentation. PoolFormer serves as corresponding teacher ANN. $\dagger$: Reproduce result from the source code of original work.\label{tab:odss}}
\resizebox{0.7\textwidth}{!}{
\begin{tabular}{lcccccc}
\toprule
\multirow{2}{*}{Method}     & \multicolumn{3}{l}{Object Detection}           & \multicolumn{3}{l}{Semantic Segmentation}   \\
  & P(M) & T & mAP(\%) & P(M) & T & MIoU(\%) \\ \midrule
PoolFormer  & 40.6        & 1 & 60.1 & 34.6        & 1 & 42.0        \\ \hline
M-Sformer & 75.0        & 1 & \cellcolor{baselinecolor}51.2 & 58.9        & 4 & \cellcolor{baselinecolor}34.2$^\dagger$       \\
\begin{tabular}[c]{@{}l@{}}BKDSNN\end{tabular}     & 75.0        & 1 & \textbf{\begin{tabular}[c]{@{}c@{}}52.3(+1.1)\end{tabular}} & 58.9        & 4 &  \textbf{\begin{tabular}[c]{@{}c@{}}35.5(+1.3)\end{tabular}} \\ \bottomrule
\end{tabular}
}
\vskip -0.2cm
\vskip -0.5cm
\end{table}

We extend BKDSNN on object detection and semantic segmentation tasks in~\cref{tab:odss} with Meta-SpikeFormer~\cite{yao2023spike}, our BKDSNN surpasses Meta-SpikeFormer on both object detection and semantic segmentation tasks, which further verifies the effectiveness of BKDSNN.
\end{document}